\documentclass[10pt,journal,cspaper,compsoc]{IEEEtran}

\usepackage{amsmath,amssymb}
\usepackage[pdftex]{graphicx}
\usepackage{hyperref}
\usepackage{algorithm} 
\usepackage{algorithmic}
\usepackage{multirow}
\usepackage{cite}
\usepackage{url}

\begin{document}

\title{Security Evaluation of Pattern Classifiers\\ under Attack}

\author{Battista~Biggio,~\IEEEmembership{Member,~IEEE,}
Giorgio~Fumera,~\IEEEmembership{Member,~IEEE,}
Fabio~Roli,~\IEEEmembership{Fellow,~IEEE}
\thanks{Please cite this work as: \emph{B. Biggio, G. Fumera, and F. Roli. Security evaluation of pattern classifiers under attack. IEEE Transactions on Knowledge and Data Engineering, 26(4):984-996, April 2014.}}
\thanks{The authors are with the Department of Electrical and Electronic Engineering, University of Cagliari, Piazza d'Armi, 09123 Cagliari, Italy}
\thanks{Battista Biggio: e-mail battista.biggio@diee.unica.it, phone +39 070 675 5776}
\thanks{Giorgio Fumera: e-mail fumera@diee.unica.it, phone +39 070 675 5754}
\thanks{Fabio Roli (corresponding author): e-mail roli@diee.unica.it, phone +39 070 675 5779, fax (shared) +39 070 675 5782}
}

\IEEEcompsoctitleabstractindextext{
\begin{abstract}
Pattern classification systems are commonly used in \emph{adversarial} applications, like biometric authentication, network intrusion detection, and spam filtering, in which data can be purposely manipulated by humans to undermine their operation.
As this adversarial scenario is not taken into account by classical design methods, pattern classification systems may exhibit vulnerabilities, whose exploitation may severely affect their performance, and consequently limit their practical utility.
Extending pattern classification theory and design methods to adversarial settings is thus a novel and very relevant research direction, which has not yet been pursued in a systematic way.
In this paper, we address one of the main open issues: evaluating at design phase the \emph{security} of pattern classifiers, namely, the performance degradation under potential attacks they may incur during operation.
We propose a framework for \emph{empirical} evaluation of classifier security that formalizes and generalizes the main ideas proposed in the literature, and give examples of its use in three real applications.
Reported results show that security evaluation can provide a more complete understanding of the classifier's behavior in adversarial environments, and lead to better design choices.
\end{abstract}

\begin{IEEEkeywords}
Pattern classification, adversarial classification, performance evaluation, security evaluation, robustness evaluation.
\end{IEEEkeywords}}

\maketitle

\IEEEdisplaynotcompsoctitleabstractindextext

\section{Introduction}
\label{sect:introduction}

Pattern classification systems based on machine learning algorithms are commonly used in security-related applications like biometric authentication, network intrusion detection, and spam filtering, to discriminate between a ``legitimate'' and a ``malicious'' pattern class (e.g., legitimate and spam emails).
Contrary to traditional ones, these applications have an intrinsic adversarial nature since the input data can be purposely manipulated by an intelligent and adaptive \emph{adversary} to undermine classifier operation.
This often gives rise to an arms race between the adversary and the classifier designer.
Well known examples of attacks against pattern classifiers are:
submitting a fake biometric trait to a biometric authentication system (\emph{spoofing} attack) \cite{rodrigues09,johnson10}; modifying network packets belonging to intrusive traffic to evade intrusion detection systems \cite{fogla06}; manipulating the content of spam emails to get them past spam filters (e.g., by misspelling common spam words to avoid their detection) \cite{wittel04,lowd05-ceas,kolcz09}.
Adversarial scenarios can also occur in intelligent data analysis \cite{skillicorn09} and information retrieval \cite{fetterly07}; e.g., a malicious webmaster may manipulate search engine rankings to artificially promote her\footnote{The adversary is hereafter referred to as feminine due to the popular interpretation as ``Eve'' or ``Carol'' in cryptography and computer security.} web site.

It is now acknowledged that, since pattern classification systems based on classical theory and design methods \cite{duda-hart-stork} do not take into account adversarial settings, they exhibit vulnerabilities to several potential attacks, allowing adversaries to undermine their effectiveness \cite{dalvi04,barreno-ASIACCS06,cardenas-ws06,kolcz09,laskov10-ed,huang11}.
A systematic and unified treatment of this issue is thus needed to allow the trusted adoption of pattern classifiers in adversarial environments, starting from the theoretical foundations up to novel design methods, extending the classical design cycle of \cite{duda-hart-stork}.
In particular, three main open issues can be identified:
(i) analyzing the vulnerabilities of classification algorithms, and the corresponding attacks \cite{barreno-ASIACCS06,barreno10,huang11};
(ii) developing novel methods to assess classifier security against these attacks, which is not possible using classical performance evaluation methods \cite{lowd05,cardenas-ws06,kolcz09,laskov09};
(iii) developing novel design methods to guarantee classifier security in adversarial environments \cite{dalvi04,kolcz09,rodrigues09}.

Although this emerging field is attracting growing interest \cite{nips07-adv,dagstuhl12-adv,laskov10-ed}, the above issues have only been sparsely addressed under different perspectives and to a limited extent. 
Most of the work has focused on application-specific issues related to spam filtering and network intrusion detection, e.g., \cite{dalvi04,wittel04,lowd05-ceas,kolcz09,fogla06}, while 
only a few theoretical models of adversarial classification problems have been proposed in the machine learning literature \cite{dalvi04,barreno10,huang11}; however, they do not yet provide practical guidelines and tools for designers of pattern recognition systems.

Besides introducing these issues to the pattern recognition research community, in this work we address issues (i) and (ii) above by developing a framework for the \emph{empirical} evaluation of classifier security at design phase that extends the model selection and performance evaluation steps of the classical design cycle of \cite{duda-hart-stork}.

In Sect.~\ref{sect:background} we summarize previous work, and point out three main ideas that emerge from it.
We then formalize and generalize them in our framework (Sect.~\ref{sect:framework}).
First, to pursue security in the context of an arms race it is not sufficient to \emph{react} to observed attacks, but it is also necessary to \emph{proactively anticipate} the adversary by \emph{predicting} the most relevant, potential attacks through a what-if analysis;
this allows one to develop suitable countermeasures \emph{before} the attack actually occurs, according to the principle of \emph{security by design}.
Second, to provide practical guidelines for simulating realistic attack scenarios, we define a general model of the adversary, in terms of her goal, knowledge, and capability, which encompasses and generalizes models proposed in previous work.
Third, since the presence of carefully targeted attacks may affect the distribution of training and testing data separately, we propose a model of the data distribution that can formally characterize this behavior, and that allows us to take into account a large number of potential attacks; we also propose an algorithm for the generation of training and testing sets to be used for security evaluation, which can naturally accommodate application-specific and heuristic techniques for simulating attacks.

In Sect.~\ref{sect:experiments} we give three concrete examples of applications of our framework in spam filtering, biometric authentication, and network intrusion detection.
In Sect.~\ref{sect:secure-design}, we discuss how the classical design cycle of pattern classifiers should be revised to take security into account.
Finally, in Sect.~\ref{sect:open-issues}, we summarize our contributions, the limitations of our framework, and some open issues.

\section{Background and previous work}
\label{sect:background}

Here we review previous work, highlighting the concepts that will be exploited in our framework.

\subsection{A taxonomy of attacks against pattern classifiers}
\label{sect:background-adversarial-barreno}

A taxonomy of potential attacks against pattern classifiers was proposed in \cite{barreno-ASIACCS06,barreno10}, and subsequently extended in \cite{huang11}. We will exploit it in our framework, as part of the definition of attack scenarios.
The taxonomy is based on two main features: the kind of \emph{influence} of attacks on the classifier, and the kind of \emph{security violation} they cause.
The influence can be either {\bf causative}, if it undermines the learning algorithm to cause subsequent misclassifications;
or {\bf exploratory}, if it exploits knowledge of the trained classifier to cause misclassifications, without affecting the learning algorithm. Thus, causative attacks may influence both training and testing data, or only training data, whereas exploratory attacks affect only testing data.
The security violation can be an {\bf integrity} violation, if it allows the adversary to access the service or resource protected by the classifier;
an {\bf availability} violation, if it denies legitimate users access to it; or a {\bf privacy} violation, if it allows the adversary to obtain confidential information from the classifier.
Integrity violations result in misclassifying malicious samples as legitimate, while availability violations can also cause legitimate samples to be misclassified as malicious.
A third feature of the taxonomy is the \emph{specificity} of an attack, that ranges from {\bf targeted} to {\bf indiscriminate}, depending on whether the attack focuses on a single or few specific samples (e.g., a specific spam email misclassified as legitimate), or on a wider set of samples.

\subsection{Limitations of classical performance evaluation methods in adversarial classification}
\label{sect:background-traditional}

Classical performance evaluation methods, like $k$-fold cross validation and bootstrapping, aim to estimate the performance that a classifier will exhibit during \emph{operation}, by using data $\mathcal D$ collected during classifier \emph{design}.\footnote{By \emph{design} we refer to the classical steps of \cite{duda-hart-stork}, which include feature extraction, model selection, classifier training and performance evaluation (testing). \emph{Operation} denotes instead the phase which starts when the deployed classifier is set to operate in a real environment.}
These methods are based on the \emph{stationarity} assumption that the data seen during operation follow the same distribution as $\mathcal D$. Accordingly, they resample $\mathcal D$ to construct one or more pairs of training and testing sets that ideally follow the same distribution as $\mathcal D$ \cite{duda-hart-stork}.
However, the presence of an intelligent and adaptive adversary makes the classification problem highly non-stationary, and makes it difficult to predict how many and which kinds of attacks a classifier will be subject to during operation, that is, how the data distribution will change.
In particular, the testing data processed by the trained classifier can be affected by both exploratory and causative attacks, while the training data can only be affected by causative attacks, if the classifier is retrained online \cite{barreno-ASIACCS06,barreno10,huang11}.\footnote{``Training'' data refers both to the data used by the learning algorithm during classifier design, coming from $\mathcal D$, and to the data collected during operation to retrain the classifier through online learning algorithms. ``Testing'' data refers both to the data drawn from $\mathcal D$ to evaluate classifier performance during design, and to the data classified during operation. The meaning will be clear from the context.}
In both cases, during operation, testing data may follow a \emph{different} distribution than that of training data, when the classifier is under attack. Therefore, security evaluation can not be carried out according to the classical paradigm of performance evaluation.\footnote{Classical performance evaluation is generally not suitable for non-stationary, time-varying environments, where changes in the data distribution can be hardly predicted; e.g., in the case of \emph{concept drift} \cite{kuncheva07}. Further, the evaluation process has a different goal in this case, i.e., to assess whether the classifier can ``recover'' quickly \emph{after} a change has occurred in the data distribution. Instead, security evaluation aims at identifying the most relevant attacks and threats that should be countered \emph{before} deploying the classifier (see Sect.~\ref{sect:background-arms-race}).}

\subsection{Arms race and security by design}
\label{sect:background-arms-race}

\begin{figure*}[ht]
\begin{center}
\includegraphics[width=0.99\textwidth]{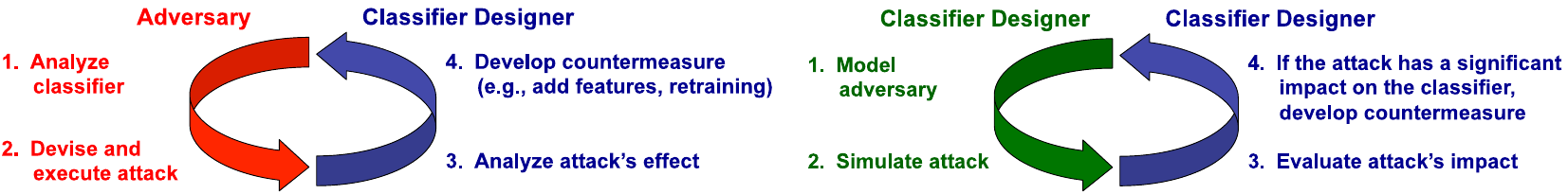}
\caption{A conceptual representation of the \emph{arms race} in adversarial classification. \emph{Left:} the classical ``reactive'' arms race. The  designer reacts to the attack by analyzing the attack's effects and developing countermeasures. \emph{Right:} the ``proactive'' arms race advocated in this paper. The designer tries to anticipate the adversary by simulating potential attacks, evaluating their effects, and developing countermeasures if necessary.}
\label{fig:arms-race}
\end{center}
\end{figure*}

Security problems often lead to a ``reactive'' arms race between the adversary and the classifier designer. At each step, the adversary analyzes the classifier defenses, and develops an attack strategy to overcome them. The designer reacts by analyzing the novel attack samples, and, if required, updates the classifier; typically, by retraining it on the new collected samples, and/or adding features that can detect the novel attacks (see Fig.~\ref{fig:arms-race}, \emph{left}).
Examples of this arms race can be observed in spam filtering and malware detection, where it has led to a considerable increase in the variability and sophistication of attacks and countermeasures. 

To secure a system, a common approach used in engineering and cryptography is \emph{security by obscurity}, that relies on keeping secret some of the system details to the adversary. In contrast, the paradigm of \emph{security by design} advocates that systems should be designed from the ground-up to be secure, without assuming that the adversary may ever find out some important system details. Accordingly, the system designer should \emph{anticipate} the adversary by simulating a ``proactive'' arms race to (i) figure out the most relevant threats and attacks, and (ii) devise proper countermeasures, \emph{before} deploying the classifier (see Fig.~\ref{fig:arms-race}, \emph{right}).
This paradigm typically improves security by delaying each step of the ``reactive'' arms race, as it requires the adversary to spend a greater effort (time, skills, and resources) to find and exploit vulnerabilities.
System security should thus be guaranteed for a longer time, with less frequent supervision or human intervention.

The goal of security evaluation is to address issue (i) above, i.e., to \emph{simulate} a number of realistic attack scenarios that may be incurred during operation, and to assess the impact of the corresponding attacks on the targeted classifier to highlight the most critical vulnerabilities. This amounts to performing a \emph{what-if} analysis \cite{rizzi09}, which is a common practice in security. This approach has been implicitly followed in several previous works (see Sect.~\ref{sect:background-adversarial}), but never formalized within a general framework for the empirical evaluation of classifier security.
Although security evaluation may also suggest specific countermeasures, the design of \emph{secure} classifiers, i.e., issue (ii) above, remains a distinct open problem.

\subsection{Previous work on security evaluation}
\label{sect:background-adversarial}

Many authors implicitly performed security evaluation as a what-if analysis, based on empirical simulation methods; however, they mainly focused on a specific application, classifier and attack, and devised ad hoc security evaluation procedures based on the exploitation of problem knowledge and heuristic techniques \cite{wittel04,lowd05-ceas,newsome06,globerson-ICML06,fogla06,perdisci-ICDM06,chung07,jorgensen08,cretu08,nelson08,rodrigues09,kolcz09,rubinstein09,kloft10,dekel10,barreno10,johnson10,biggio11-smc,biggio-IJMLC10,biggio11-mcs,biggio12-spr,biggio12-icml}.
Their goal was either to point out a previously unknown vulnerability, or to evaluate security against a known attack. In some cases, specific countermeasures were also proposed, according to a proactive/security-by-design approach.
Attacks were simulated by manipulating training and testing samples according to application-specific criteria only, without reference to more general guidelines; consequently, such techniques can not be directly exploited by a system designer in more general cases.

A few works proposed \emph{analytical} methods to evaluate the security of learning algorithms or of some classes of decision functions (e.g., linear ones), based on more general, application-independent criteria to model the adversary's behavior (including PAC learning and game theory) \cite{kearns93,dalvi04,lowd05,cardenas06,cardenas-ws06,biggio09-MCS,laskov09,huang11,brueckner12}.
Some of these criteria will be exploited in our framework for \emph{empirical} security evaluation; in particular, in the definition of the adversary model described in Sect.~\ref{sect:framework-adversary}, as high-level guidelines for simulating attacks.

\subsection{Building on previous work}
\label{sect:background-contribution}

We summarize here the three main concepts more or less explicitly emerged from previous work that will be exploited in our framework for security evaluation.

\textbf{1) Arms race and security by design}: since it is not possible to predict how many and which kinds of attacks a classifier will incur during operation, classifier security should be proactively evaluated using a what-if analysis, by \emph{simulating} potential attack scenarios.

\textbf{2) Adversary modeling}: effective simulation of attack scenarios requires a formal model of the adversary.

\textbf{3) Data distribution under attack}: the distribution of testing data may differ from that of training data, when the classifier is under attack.

\section{A framework for empirical evaluation of classifier security}
\label{sect:framework}

We propose here a framework for the empirical evaluation of classifier security in adversarial environments, that unifies and builds on the three concepts highlighted in Sect.~\ref{sect:background-contribution}.
Our main goal is to provide a quantitative and general-purpose basis for the application of the \emph{what-if analysis} to classifier security evaluation, based on the definition of potential attack scenarios.
To this end, we propose: (i) a model of the adversary, that allows us to define any attack scenario;
(ii) a corresponding model of the data distribution;
and (iii) a method for generating training and testing sets that are representative of the data distribution, and are used for empirical performance evaluation.

\subsection{Attack scenario and model of the adversary}
\label{sect:framework-adversary}

Although the definition of attack scenarios is ultimately an application-specific issue, it is possible to give general guidelines that can help the designer of a pattern recognition system.
Here we propose to specify the attack scenario in terms of a conceptual model of the adversary that encompasses, unifies, and extends different ideas from previous work.
Our model is based on the assumption that the adversary acts \emph{rationally} to attain a given \emph{goal}, according to her \emph{knowledge} of the classifier, and her \emph{capability} of manipulating data. This allows one to derive the corresponding optimal \emph{attack strategy}.

\textbf{Adversary's goal}. As in \cite{laskov09}, it is formulated as the optimization of an objective function.
We propose to define this function based on the desired security violation (\emph{integrity}, \emph{availability}, or \emph{privacy}), and on the attack \emph{specificity} (from \emph{targeted} to \emph{indiscriminate}), according to the taxonomy in \cite{barreno-ASIACCS06,huang11} (see Sect.~\ref{sect:background-adversarial-barreno}).
For instance, the goal of an indiscriminate integrity violation may be to maximize the fraction of misclassified malicious samples \cite{dalvi04,kolcz09,huang11};
the goal of a targeted privacy violation may be to obtain some specific, confidential information from the classifier (e.g., the biometric trait of a given user enrolled in a biometric system) by exploiting the class labels assigned to some ``query'' samples, while minimizing the number of query samples that the adversary has to issue to violate privacy \cite{adler05,lowd05,huang11}.

\textbf{Adversary's knowledge}.
Assumptions on the adversary's knowledge have only been qualitatively discussed in previous work, mainly depending on the application at hand.
Here we propose a more systematic scheme for their definition, with respect to the knowledge of the single components of a pattern classifier:
(k.i) the training data;
(k.ii) the feature set;
(k.iii) the learning algorithm and the kind of decision function (e.g., a linear SVM);
(k.iv) the classifier's decision function and its parameters (e.g., the feature weights of a linear classifier);
(k.v) the feedback available from the classifier, if any (e.g., the class labels assigned to some ``query'' samples that the adversary issues to get feedback \cite{adler05,lowd05,huang11}). 
It is worth noting that realistic and minimal assumptions about what can be kept fully secret from the adversary should be done, as discussed in \cite{huang11}. Examples of adversary's knowledge are given in Sect.~\ref{sect:experiments}.

\textbf{Adversary's capability}. It refers to the control that the adversary has on training and testing data.
We propose to define it in terms of:
(c.i) the attack influence (either \emph{causative} or \emph{exploratory}), as defined in \cite{barreno-ASIACCS06,huang11};
(c.ii) whether and to what extent the attack affects the class priors;
(c.iii) how many and which training and testing samples can be controlled by the adversary in each class;
(c.iv) which features can be manipulated, and to what extent, taking into account application-specific constraints (e.g., correlated features can not be modified independently, and the functionality of malicious samples can not be compromised \cite{dalvi04,kolcz09,fogla06,rodrigues09}).

\textbf{Attack strategy}. One can finally define the \emph{optimal} attack strategy, namely, how training and testing data should be \emph{quantitatively} modified to optimize the objective function characterizing the adversary's goal. 
Such modifications are defined in terms of:
(a.i) how the class priors are modified;
(a.ii) what fraction of samples of each class is affected by the attack;
and (a.iii) how features are manipulated by the attack.
Detailed examples are given in Sect.~\ref{sect:experiments}.

Once the \emph{attack scenario} is defined in terms of the adversary model and the resulting attack strategy, our framework proceeds with the definition of the corresponding data distribution, that is used to construct training and testing sets for security evaluation.

\subsection{A model of the data distribution}
\label{sect:framework-model}

We consider the standard setting for classifier design in a problem which consists of discriminating between legitimate (L) and malicious (M) samples: a learning algorithm and a performance measure have been chosen, a set $\mathcal D$ of $n$ labelled samples has been collected, and a set of $d$ features have been extracted, so that $\mathcal D=\{(\mathbf x_i, y_i)\}_{i=1}^n$, where
$\mathbf x_{i}$ denotes a $d$-dimensional feature vector,
and $y_{i} \in \{{\rm L},{\rm M}\}$ a class label.
The pairs $(\mathbf x_i, y_i)$ are assumed to be i.i.d.~samples of some unknown distribution $p_{\mathcal D}(\mathbf X, Y)$.
Since the adversary model in Sect.~\ref{sect:framework-adversary} requires us to specify how the attack affects the class priors and the features of each class, we consider the classical generative model $p_{\mathcal D}(\mathbf X,Y)=p_{\mathcal D}(Y)p_{\mathcal D}(\mathbf X | Y)$.\footnote{In this paper, for the sake of simplicity, we use the lowercase notation $p(\cdot)$ to denote any probability density function, with both continuous and discrete random variables.}
To account for the presence of attacks during operation, which may affect either the training or the testing data, or both, we denote the corresponding training and testing distributions as $p_{\rm tr}$ and $p_{\rm ts}$, respectively. We will just write $p$ when we want to refer to either of them, or both, and the meaning is clear from the context.

When a classifier is not under attack, according to the classical stationarity assumption we have
$p_{\rm tr}(Y) = p_{\rm ts}(Y) = p_{\mathcal D}(Y)$, and
$p_{\rm tr}(\mathbf X | Y) = p_{\rm ts}(\mathbf X | Y) = p_{\mathcal D}(\mathbf X | Y)$.
We extend this assumption to the components of $p_{\rm tr}$ and $p_{\rm ts}$ that are not affected by the attack (if any), by assuming that they remain identical to the corresponding distribution $p_{\mathcal D}$ (e.g., if the attack does not affect the class priors, the above equality also holds under attack).

The distributions $p(Y)$ and $p(\mathbf X | Y)$ that are affected by the attack can be defined as follows, according to the definition of the attack strategy, (a.i-iii).

\textbf{Class priors}. $p_{\rm tr}(Y)$ and $p_{\rm ts}(Y)$ can be immediately defined based on assumption (a.i).

\textbf{Class-conditional distributions}. $p_{\rm tr}(\mathbf X | Y)$ and $p_{\rm ts}(\mathbf X | Y)$ can be defined based on assumptions (a.ii-iii). First, to account for the fact that the attack may not modify all training or testing samples, according to (a.ii), we model $p(\mathbf X | Y)$ as a mixture controlled by a Boolean random variable $A$, which denotes whether a given sample has been manipulated ($A={\rm T}$) or not ($A={\rm F}$):
\begin{eqnarray}
\label{eq:data-model-mixture}
\nonumber p(\mathbf X | Y) & = & p(\mathbf X | Y, A={\rm T}) p(A={\rm T} | Y) +   \\
 & & p(\mathbf X | Y, A={\rm F}) p(A={\rm F} | Y) .
\end{eqnarray}
We name the samples of the component $p(\mathbf X | Y, A={\rm T})$ \emph{attack samples} to emphasize that their distribution is different from that of samples which have not been manipulated by the adversary, $p(\mathbf X | Y, A={\rm F})$. Note that $p(A={\rm T} | Y=y)$ is the probability that an attack sample belongs to class $y$, i.e., the percentage of samples of class $y$ controlled by the adversary. It is thus defined by (a.ii).

For samples which are unaffected by the attack, the stationarity assumption holds:
\begin{equation}
\label{eq:data-model-D}
p(\mathbf X | Y, A={\rm F}) = p_{\mathcal D}(\mathbf X | Y) .
\end{equation}

The distribution $p(\mathbf X | Y, A={\rm T})$ depends on assumption (a.iii).
Depending on the problem at hand, it may not be possible to analytically define it.
Nevertheless, for the purpose of security evaluation, we will show in Sect.~\ref{sect:framework-algorithm} that it can be defined as the \emph{empirical} distribution of a set of fictitious attack samples.

Finally, $p(\mathbf X | Y, A={\rm F})$ can be defined similarly, if its analytical definition is not possible. In particular, according to Eq.~(\ref{eq:data-model-D}), it can be defined as the empirical distribution of $\mathcal D$.

\begin{figure}[t]
\begin{center}
\includegraphics[width=0.18\textwidth]{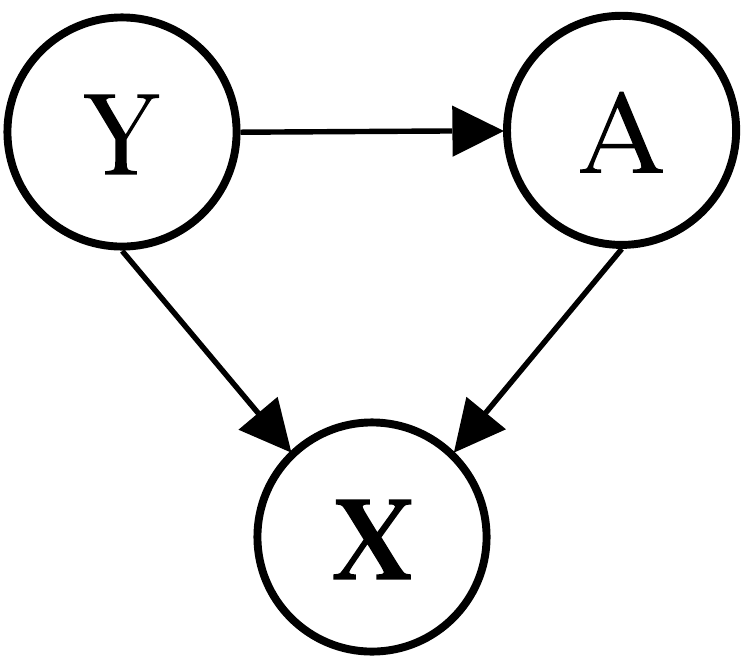}
\caption{Generative model of the distributions $p_{\rm tr}$ and $p_{\rm ts}$.}
\label{fig:graphical-model}
\end{center}
\end{figure}

The above generative model of the training and testing distributions $p_{\rm tr}$ and $p_{\rm ts}$ is represented by the Bayesian network in Fig.~\ref{fig:graphical-model}, which corresponds to factorize $p(\mathbf X,Y,A)$ as follows:
\begin{equation}
\label{eq:gen-model}
p(\mathbf X,Y,A) = p(Y) p(A | Y) p(\mathbf X | Y, A) .
\end{equation} 

Our model can be easily extended to take into account concurrent attacks involving $m>1$ different kinds of sample manipulations; for example, to model attacks against different classifiers in multimodal biometric systems \cite{rodrigues09,johnson10}.
To this end, one can define $m$ different Boolean random variables $\mathbf A = (A_1,\ldots,A_m)$, and the corresponding distributions. 
The extension of Eq.~\ref{eq:gen-model} is then straightforward (e.g., see \cite{biggio11-smc}). The dependence between the different attacks, if any, can be modeled by the distribution $p(\mathbf A | Y)$.

If one is also interested in evaluating classifier security subject to temporal, non-adversarial variations of the data distribution (e.g., the drift of the content of legitimate emails over time), it can be assumed that the distribution $p(\mathbf X,Y, A={\rm F})$ changes over time, according to some model $p(\mathbf X,Y, A={\rm F}, t)$ (see, e.g., \cite{kuncheva07}). Then, classifier security at time $t$ can be evaluated by modeling the attack distribution over time $p(\mathbf X,Y, A={\rm T}, t)$ as a function of $p(\mathbf X,Y, A={\rm F}, t)$.

The model of the adversary in Sect.~\ref{sect:framework-adversary}, and the above model of the data distribution provide a quantitative, well-grounded and general-purpose basis for the application of the what-if analysis to classifier security evaluation, which advances previous work.

\subsection{Training and testing set generation}
\label{sect:framework-algorithm}

Here we propose an algorithm to sample training (TR) and testing (TS) sets of any desired size from the distributions $p_{\rm tr}(\mathbf X,Y)$ and $p_{\rm ts}(\mathbf X,Y)$.

We assume that $k \geq 1$ different pairs of training and testing sets $(\mathcal D^{i}_{\rm TR},\mathcal D^{i}_{\rm TS})$, $i=1,\ldots,k$, have been obtained from $\mathcal D$ using a classical resampling technique, like cross-validation or bootstrapping. Accordingly, their samples follow the distribution $p_{\mathcal D}(\mathbf X,Y)$.
In the following, we describe how to modify each of the sets $\mathcal D^{i}_{\rm TR}$ to construct a training set TR$^{i}$ that follows the distribution $p_{\rm tr}(\mathbf X,Y)$. For the sake of simplicity, we will omit the superscript $i$.
An identical procedure can be followed to construct a testing set TS$^{i}$ from each of the $\mathcal D^{i}_{\rm TS}$.
Security evaluation is then carried out with the classical method, by averaging (if $k>1$) the performance of the classifier trained on TR$^{i}$ and tested on TS$^{i}$.

If the attack does not affect the training samples, i.e., $p_{\rm tr}(\mathbf X,Y)=p_{\mathcal D}(\mathbf X,Y)$, TR is simply set equal to $\mathcal D_{\rm TR}$. Otherwise, two alternatives are possible.
(i) If $p_{\rm tr}(\mathbf X | Y, A)$ is analytically defined for each $Y \in \{ {\rm L},{\rm M} \}$ and $A \in \{ {\rm T}, {\rm F} \}$, then TR can be obtained by sampling the generative model of $p(\mathbf X,Y,A)$ of Eq.~(\ref{eq:gen-model}):
first, a class label $y$ is sampled from $p_{\rm tr}(Y)$, then, a value $a$ from $p_{\rm tr}(A|Y=y)$, and, finally, a feature vector $\mathbf x$ from $p_{\rm tr}(\mathbf X | Y=y, A=a)$.
(ii) If $p_{\rm tr}(\mathbf X | Y=y, A=a)$ is not analytically defined for some $y$ and $a$, but a set of its samples is available, denoted in the following as $\mathcal D_{\rm TR}^{y,a}$, it can be approximated as the empirical distribution of  $\mathcal D_{\rm TR}^{y,a}$. Accordingly, we can sample with replacement from $\mathcal D_{\rm TR}^{y,a}$ \cite{efron93}.
An identical procedure can be used to construct the testing set TS.
The procedure to obtain TR or TS is formally described as Algorithm \ref{alg:data-set-construction}.\footnote{Since the proposed algorithm is based on classical resampling techniques such as cross-validation and bootstrapping, it is reasonable to expect that the bias and the variance of the estimated classification error (or of any other performance measure) will enjoy similar statistical properties to those exhibited by classical performance evaluation methods based on the same techniques. In practice, these error components are typically negligible with respect to errors introduced by the use of limited training data, biased learning/classification algorithms, and noisy or corrupted data.}

\begin{algorithm}[t]
\label{alg:data-set-construction}
\caption{Construction of TR or TS.}
\textbf{Input:} 
The number $n$ of desired samples; \\
the distributions $p(Y)$ and $p(A|Y)$;\\
for each $y \in \{{\rm L},{\rm M}\}, a \in \{{\rm T},{\rm F}\}$, the distribution $p(\mathbf X | Y=y, A=a)$, if analytically defined, or the set of samples $\mathcal D^{y,a}$, otherwise. \\
\textbf{Output:} A data set $\mathcal S$ (either TR or TS) 
drawn from $p(Y)p(A|Y)p(\mathbf X|Y, A)$.
\label{alg:data-set-construction}
\begin{algorithmic}[1]
\STATE{$S \gets \emptyset$}
\FOR{$i = 1,\ldots,n$}
\STATE{sample $y$ from $p(Y)$} \label{algo:step-y}
\STATE{sample $a$ from $p(A|Y=y)$} \label{algo:step-a}
\STATE{draw a sample $\mathbf x$ from $p(\mathbf X | Y=y, A=a)$, if analytically defined; otherwise, sample with replacement from $\mathcal D^{y,a}$} \label{algo:step-x}
\STATE{$\mathcal S \gets \mathcal S \bigcup \ \{ (\mathbf x, y) \}$} 
\ENDFOR
\RETURN{$\mathcal S$}
\end{algorithmic}
\end{algorithm}

Let us now discuss how to construct the sets $\mathcal D_{\rm TR}^{y,a}$, when $p_{\rm tr}(\mathbf X | Y=y, A=a)$ is not analytically defined. The same discussion holds for the sets $\mathcal D_{\rm TS}^{y,a}$. First, the two sets $\mathcal D_{\rm TR}^{{\rm L},{\rm F}}$ and $\mathcal D_{\rm TR}^{{\rm M},{\rm F}}$
can be respectively set equal to the legitimate and malicious samples in $\mathcal D_{\rm TR}$, since the distribution of such samples is assumed to be just $p_{\rm tr}(\mathbf X|Y={\rm L},A={\rm F})$ and $p_{\rm tr}(\mathbf X|Y={\rm M},A={\rm F})$ (see Eq.~\ref{eq:data-model-D}):
$\mathcal D_{\rm TR}^{{\rm L},{\rm F}} = \{ (\mathbf x, y) \in \mathcal D_{\rm TR} : y={\rm L} \}$,
$\mathcal D_{\rm TR}^{{\rm M},{\rm F}} = \{ (\mathbf x, y) \in \mathcal D_{\rm TR} : y={\rm M} \}$.
The two sets of attack samples $\mathcal D_{\rm TR}^{y,{\rm T}}$, for $y={\rm L}, {\rm M}$, must come instead from $p_{\rm tr}(\mathbf X|Y=y,A={\rm T})$.
They can thus be constructed according to the point (a.iii) of the attack strategy, using any technique for simulating attack samples.
Therefore, all the ad hoc techniques used in previous work for constructing fictitious attack samples can be used as methods to define or empirically approximate the distribution $p(\mathbf X | Y=y, A={\rm T})$, for $y={\rm L}, {\rm M}$.

Two further considerations must be made on the simulated attack samples $\mathcal D_{\rm TR}^{y,{\rm T}}$.
First, the number of distinct attack samples that can be obtained depends on the simulation technique, when it requires the use of the data in $\mathcal D_{\rm TR}$;
for instance, if it consists of modifying \emph{deterministically} each malicious sample in $\mathcal D_{\rm TR}$, no more than
$| \{ (\mathbf x, y) \in \mathcal D_{\rm TR} : y = {\rm M} \} |$
distinct attack samples can be obtained.
If any number of attack samples can be generated, instead (e.g., \cite{lowd05-ceas,nelson08,biggio11-mcs,biggio12-icml}), then the sets $\mathcal D_{\rm TR}^{y,{\rm T}}$, for $y={\rm L},{\rm M}$, do not need to be constructed beforehand: a single attack sample can be generated online at step \ref{algo:step-x} of Algorithm \ref{alg:data-set-construction}, when $a={\rm T}$.
Second, in some cases it may be necessary to construct the attack samples incrementally; for instance, in the causative attacks considered in \cite{barreno-ASIACCS06,kloft10,biggio12-icml}, the attack samples are added to the training data one at a time, since each of them is a function of the \emph{current} training set.
This corresponds to a non-i.i.d. sampling of the attack distribution. In these cases, Algorithm~\ref{alg:data-set-construction} can be modified by first generating all pairs $y,a$, then, the feature vectors $\mathbf x$ corresponding to $a={\rm F}$, and, lastly, the attack samples corresponding to $a={\rm T}$, one at a time.

\subsection{How to use our framework}
\label{sect:framework-application}

We summarize here the steps that the designer of a pattern classifier should take to evaluate its security using our framework, for each attack scenario of interest.
They extend the performance evaluation step of the classical design cycle of \cite{duda-hart-stork}, which is used as part of the model selection phase, and to evaluate the final classifier to be deployed.

\textbf{1) Attack scenario}. The attack scenario should be defined at the conceptual level by making specific assumptions on the goal, knowledge (k.i-v), and capability of the adversary (c.i-iv), 
and defining the corresponding attack strategy (a.i-iii), 
according to the model of Sect.~\ref{sect:framework-adversary}. 

\textbf{2) Data model}. According to the hypothesized attack scenario, the designer should define the distributions $p(Y)$, $p(A|Y)$, and $p(\mathbf X|Y,A)$, for $Y \in \{{\rm L}, {\rm M}\}$, $A \in \{{\rm F}, {\rm T}\}$, and for training and testing data. 
If $p(\mathbf X|Y,A)$ is not analytically defined for some $Y=y$ and $A=a$, either for training or testing data, the corresponding set $\mathcal D_{\rm TR}^{y,a}$ or $\mathcal D_{\rm TS}^{y,a}$ must be constructed.
The sets $\mathcal D_{\rm TR}^{y,{\rm F}}$ ($\mathcal D_{\rm TS}^{y,{\rm F}}$) are obtained from $\mathcal D_{\rm TR}$ ($\mathcal D_{\rm TS}$).
The sets $\mathcal D_{\rm TR}^{y,{\rm T}}$ and $\mathcal D_{\rm TS}^{y,{\rm T}}$ can be generated, only if the attack involves sample manipulation, using an attack sample simulation technique according to the attack strategy (a.iii).

\textbf{3) Construction of TR and TS}.
Given $k\geq 1$ pairs $(\mathcal D^{i}_{\rm TR},\mathcal D^{i}_{\rm TS})$, $i=1,\ldots,k$, obtained from classical resampling techniques like cross-validation or bootstrapping, the size of TR and TS must be defined, and Algorithm \ref{alg:data-set-construction} must be run with the corresponding inputs to obtain TR$^{i}$ and TS$^{i}$. 
If the attack does not affect the training (testing) data, TR$^{i}$ (TS$^{i}$) is set to $\mathcal D^{i}_{\rm TR}$ ($\mathcal D^{i}_{\rm TS}$).

\textbf{4) Performance evaluation}. 
The classifier performance under the simulated attack is evaluated using the constructed (TR$^{i}$, TS$^{i}$) pairs, as in classical techniques.

\section{Application examples}
\label{sect:experiments}

While previous work focused on a single application, we consider here three different application examples of our framework in spam filtering, biometric authentication, and network intrusion detection.
Our aim is to show how the designer of a pattern classifier can use our framework, and what kind of additional information he can obtain from security evaluation. We will show that a trade-off between classifier accuracy and security emerges sometimes, and that this information can be exploited for several purposes;
e.g., to improve the model selection phase by considering both classification accuracy and security.

\begin{table}[t]
\begin{center}
\begin{tabular}{l|p{1.49 cm}|p{1.49 cm}|p{1.51 cm}|}
\cline{2-4}
   & \textbf{Sect.~4.1} & \textbf{Sect.~4.2} & \textbf{Sect.~4.3} \\ \hline
 \multicolumn{1}{|p{2 cm}|}{\textit{Attack scenario}} &{Indiscrim. Exploratory Integrity} &{Targeted Exploratory Integrity} &{Indiscrim. Causative Integrity} \\ \hline \hline
 \multicolumn{1}{|l|}{$p_{\rm tr}(Y)$} & $p_{\mathcal D}(Y)$ & $p_{\mathcal D}(Y)$ & $p_{\rm tr}($M$)$=$p_{\rm max}$ \\ \hline
 \multicolumn{1}{|l|}{$p_{\rm tr}(A={\rm T} | Y={\rm L})$} & 0 & 0 & 0 \\ 
\multicolumn{1}{|l|}{$p_{\rm tr}(A={\rm T} | Y={\rm M})$} & 0 & 0 & 1 \\ \hline
\multicolumn{1}{|l|}{$p_{\rm tr}(\mathbf X | Y={\rm L}, A={\rm T})$} & - & - & - \\
\multicolumn{1}{|l|}{$p_{\rm tr}(\mathbf X | Y={\rm M}, A={\rm T})$} & - & - & empirical \\ \hline
\multicolumn{1}{|l|}{$p_{\rm tr}(\mathbf X | Y={\rm L}, A={\rm F})$} & $p_{\mathcal D}(\mathbf X | {\rm L})$ & $p_{\mathcal D}(\mathbf X | {\rm L})$ & $p_{\mathcal D}(\mathbf X | {\rm L})$ \\
\multicolumn{1}{|l|}{$p_{\rm tr}(\mathbf X | Y={\rm M}, A={\rm F})$} & $p_{\mathcal D}(\mathbf X | {\rm M})$ & $p_{\mathcal D}(\mathbf X | {\rm M})$ & - \\ 
\hline \hline
\multicolumn{1}{|l|}{$p_{\rm ts}(Y)$} & $p_{\mathcal D}(Y)$ & $p_{\mathcal D}(Y)$ & $p_{\mathcal D}(Y)$ \\ \hline
 \multicolumn{1}{|l|}{$p_{\rm ts}(A={\rm T} | Y={\rm L})$} & 0 & 0 & 0 \\ 
\multicolumn{1}{|l|}{$p_{\rm ts}(A={\rm T} | Y={\rm M})$} & 1 & 1 & 0 \\\hline
\multicolumn{1}{|l|}{$p_{\rm ts}(\mathbf X | Y={\rm L}, A={\rm T})$} & - & - & -  \\
\multicolumn{1}{|l|}{$p_{\rm ts}(\mathbf X | Y={\rm M}, A={\rm T})$} & empirical & empirical & -  \\ \hline
\multicolumn{1}{|l|}{$p_{\rm ts}(\mathbf X | Y={\rm L}, A={\rm F})$} & $p_{\mathcal D}(\mathbf X | {\rm L})$ & $p_{\mathcal D}(\mathbf X | {\rm L})$ & $p_{\mathcal D}(\mathbf X | {\rm L})$ \\
\multicolumn{1}{|l|}{$p_{\rm ts}(\mathbf X | Y={\rm M}, A={\rm F})$} & - & - & $p_{\mathcal D}(\mathbf X | {\rm M})$ \\
\hline
\end{tabular}
\end{center}
\caption{Parameters of the attack scenario and of the data model for each application example. `Empirical' means that $p(\mathbf X|Y=y,A=a)$ was approximated as the empirical distribution of a set of samples $D^{y,a}$; and `-' means that no samples from this distribution were needed to simulate the attack.}
\label{table:exp}
\end{table}

\subsection{Spam filtering}
\label{sect:exp-spam}

Assume that a classifier has to discriminate between legitimate and spam emails on the basis of their textual content, and that the \emph{bag-of-words} feature representation has been chosen, with binary features denoting the occurrence of a given set of words. This kind of classifier has been considered by several authors \cite{drucker99,nelson08,kolcz09}, and it is included in several real spam filters.\footnote{SpamAssassin, \url{http://spamassassin.apache.org/}; Bogofilter, \url{http://bogofilter.sourceforge.net/}; SpamBayes \url{http://spambayes.sourceforge.net/}}

In this example, we focus on model selection.
We assume that the designer wants to choose between a support vector machine (SVM) with a linear kernel, and a logistic regression (LR) linear classifier. He also wants to choose a feature subset, among all the words occurring in training emails. A set $\mathcal D$ of legitimate and spam emails is available for this purpose. We assume that the designer wants to evaluate not only classifier accuracy in the absence of attacks, as in the classical design scenario, but also its security against the well-known bad word obfuscation (BWO) and good word insertion (GWI) attacks. They consist of modifying spam emails by inserting ``good words'' that are likely to appear in legitimate emails, and by obfuscating ``bad words'' that are typically present in spam \cite{kolcz09}. The attack scenario can be modeled as follows.

\textbf{1) Attack scenario}. \textit{Goal}.
The adversary aims at maximizing the percentage of spam emails misclassified as legitimate, which is an indiscriminate integrity violation.

\textit{Knowledge}. As in \cite{dalvi04,kolcz09}, the adversary is assumed to have perfect knowledge of the classifier, i.e.: (k.ii) the feature set, (k.iii) the kind of decision function, and (k.iv) its parameters (the weight assigned to each feature, and the decision threshold).
Assumptions on the knowledge of (k.i) the training data and (k.v) feedback from the classifier are not relevant in this case, as they do not provide any additional information.

\textit{Capability}. We assume that the adversary: 
(c.i) is only able to influence testing data (exploratory attack); 
(c.ii) can not modify the class priors;
(c.iii) can manipulate each malicious sample, but no legitimate ones;
(c.iv) can manipulate any feature value (i.e., she can insert or obfuscate any word), but up to a maximum number $n_{\rm max}$ of features in each spam email \cite{dalvi04,kolcz09}.
This allows us to evaluate how gracefully the classifier performance degrades as an increasing number of features is modified, by repeating the evaluation for increasing values of $n_{\rm max}$.

\textit{Attack strategy}.
Without loss of generality, let us further assume that $\mathbf x$ is classified as legitimate if $g(\mathbf x) = \sum_{i=1}^n w_i x_i + w_{0} < 0$, where $g(\cdot)$ is the discriminant function of the classifier, $n$ is the feature set size, $x_i \in \{0, 1\}$ are the feature values (1 and 0 denote respectively the presence and the absence of the corresponding term), $w_{i}$ are the feature weights, and $w_{0}$ is the bias.

Under the above assumptions, the optimal attack strategy can be attained by: 
(a.i) leaving the class priors unchanged;
(a.ii) manipulating \emph{all} testing spam emails;
and (a.iii) modifying up to $n_{\rm max}$ words in each spam to \emph{minimize} the discriminant function of the classifier.

Each attack sample (i.e., modified spam) can be thus obtained by solving a constrained optimization problem.
As in \cite{dalvi04}, the generation of attack samples can be represented by a function $\mathcal A : \mathcal X \mapsto \mathcal X$, and the number of modified words can be evaluated by the Hamming distance. Accordingly, for any given $\mathbf x \in \mathcal X$, the optimal attack strategy amounts to finding the attack sample $\mathcal A(\mathbf x)$ which minimizes $g(\mathcal A(\mathbf x))$, subject to the constraint that the Hamming distance between $\mathbf x$ and $\mathcal A(\mathbf x)$ is no greater than $n_{\rm max}$, that is:
\begin{eqnarray}
\nonumber  \mathcal A(\mathbf x) = & {\rm argmin}_{\mathbf{x}^{\prime}} & \sum_{i=1}^n w_i x^{\prime}_i  \\
& {\rm s.t. } & \sum_{i=1}^n | x^{\prime}_i - x_{i}| \leq n_{\rm max} . \label{eq:opt-problem-spam}
\end{eqnarray}

The solution to problem (\ref{eq:opt-problem-spam}) is straightforward. First, note that inserting (obfuscating) a word results in switching the corresponding feature value from 0 to 1 (1 to 0), and that the minimum of $g(\cdot)$ is attained when all features that have been assigned a negative (positive) weight are equal to 1 (0). 
Accordingly, for a given $\mathbf x$, the largest \emph{decrease} of $g(\mathcal A(\mathbf x))$ subject to the above constraint is obtained by analyzing all features for decreasing values of $|w_i|$, and switching from 0 to 1 (from 1 to 0) the ones corresponding to $w_i<0$ ($w_i>0$), until $n_{\rm max}$ changes have been made, or all features have been analyzed.
Note that this attack can be simulated by directly manipulating the feature vectors of spam emails instead of the emails themselves.

\textbf{2) Data model}.
Since the adversary can only manipulate testing data, we set
$p_{\rm tr}(Y) = p_{\mathcal D}(Y)$, and
$p_{\rm tr}(\mathbf X | Y) = p_{\mathcal D}(\mathbf X | Y)$.
Assumptions (a.i-ii) are directly encoded as:
(a.i) $p_{\rm ts}(Y)=p_{\mathcal D}(Y)$; and (a.ii) $p_{\rm ts}(A={\rm T} | Y={\rm M}) = 1$, and $p_{\rm ts}(A={\rm T} | Y={\rm L})=0$.
The latter implies that $p_{\rm ts}(\mathbf X | Y={\rm L})=p_{\mathcal D}(\mathbf X | Y={\rm L})$.
Assumption (a.iii) is encoded into the above function $\mathcal A(\mathbf x)$, which \emph{empirically} defines the distribution of the attack samples $p_{\rm ts}(\mathbf X | Y={\rm M}, A={\rm T})$.
Similarly, $p(\mathbf X | Y=y, A={\rm F})=p_{\mathcal D}(\mathbf X | y)$, for $y={\rm L}, {\rm M}$, will be approximated as the empirical distribution of the set $\mathcal D_{\rm TS}^{y,{\rm F}}$ obtained from $\mathcal D_{\rm TS}$, as described below; the same approximation will be made in all the considered application examples.

The definition of the attack scenario and the data model is summarized in Table~\ref{table:exp} (first column).

\textbf{3) Construction of TR and TS}.
We use a publicly available email corpus, TREC 2007. It consists of 25,220 legitimate and 50,199 real spam emails.\footnote{\url{http://plg.uwaterloo.ca/~gvcormac/treccorpus07}} We select the first 20,000 emails in chronological order to create the data set $\mathcal D$.
We then split $\mathcal D$ into two subsets $\mathcal D_{\rm TR}$ and $\mathcal D_{\rm TS}$, respectively made up of the first 10,000 and the next 10,000 emails, in chronological order. We use $\mathcal D_{\rm TR}$ to construct TR and $\mathcal D_{\rm TS}$ to construct TS, as in \cite{lowd05-ceas,kolcz09}.
Since the considered attack does not affect training samples, we set
${\rm TR}=\mathcal D_{\rm TR}$.
Then, we define $\mathcal D_{\rm TS}^{\rm L,F}$ and $\mathcal D_{\rm TS}^{\rm M,F}$ respectively as the legitimate and malicious samples in $\mathcal D_{\rm TS}$, and construct the set of attack samples $\mathcal D_{\rm TS}^{\rm M,T}$ by modifying all the samples in $\mathcal D_{\rm TS}^{\rm M,F}$ according to $\mathcal A(\mathbf x)$, for any fixed $n_{\rm max}$ value.
Since the attack does not affect the legitimate samples, we set $\mathcal D_{\rm TS}^{\rm L,T} = \emptyset$.
Finally, we set the size of TS to 10,000, and generate TS by running Algorithm \ref{alg:data-set-construction} on $\mathcal D_{\rm TS}^{\rm L,F}$, $\mathcal D_{\rm TS}^{\rm M,F}$ and $\mathcal D_{\rm TS}^{\rm M,T}$.

The features (words) are extracted from TR using the SpamAssassin tokenization method. Four feature subsets with size 1,000, 2,000, 10,000 and 20,000 have been selected using the information gain criterion \cite{sebastiani02}.

\textbf{4) Performance evaluation.}
The performance measure we use is the area under the receiver operating characteristic curve (AUC) corresponding to false positive (FP) error rates in the range $[0,0.1]$ \cite{kolcz09}:
${\rm AUC}_{10\%} = \int_0^{0.1} \rm{TP}(\rm{FP}) {\rm d} \rm{FP} \in [0,0.1]$, where TP is the true positive error rate.
It is suited to classification tasks like spam filtering, where FP errors (i.e., legitimate emails misclassified as spam) are much more harmful than false negative (FN) ones.

Under the above model selection setting (two classifiers, and four feature subsets) eight different classifier models must be evaluated.
Each model is trained on TR. SVMs are implemented with the LibSVM software \cite{libsvm}. The $C$ parameter of their learning algorithm is chosen by maximizing the AUC$_{10\%}$ through a $5$-fold cross-validation on TR. An online gradient descent algorithm is used for LR.
After classifier training, the AUC$_{10\%}$ value is assessed on TS, for different values of $n_{\rm max}$. In this case, it is a monotonically decreasing function of $n_{\rm max}$. The more graceful its decrease, the more robust the classifier is to the considered attack.
Note that, for $n_{\rm max}=0$, no attack samples are included in the testing set: the corresponding AUC$_{10\%}$ value equals that attained by classical performance evaluation methods.

\begin{figure}[t]
\begin{center}
\includegraphics[width=0.4\textwidth]{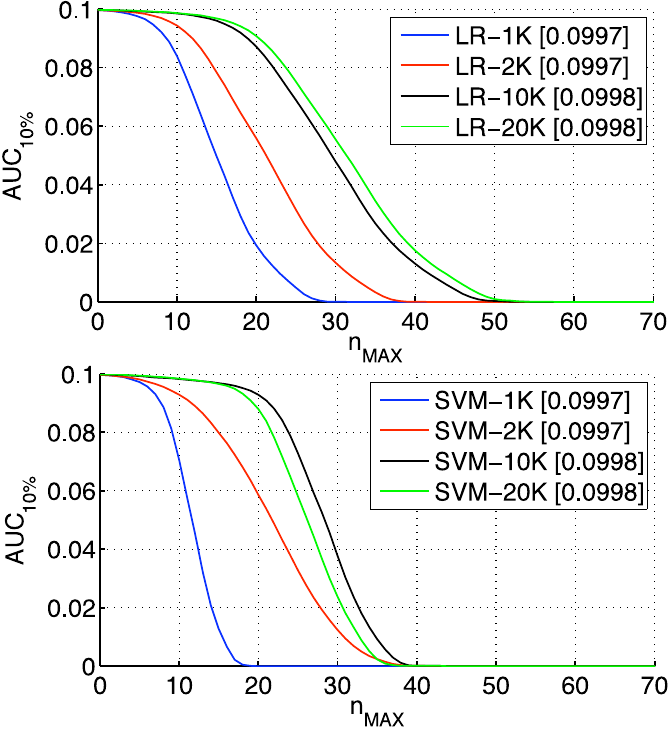}
\caption{$\rm{AUC}_{10\%}$ attained on TS as a function of $n_{\rm max}$, for the LR (top) and SVM (bottom) classifier, with 1,000 (1K), 2,000 (2K), 10,000 (10K) and 20,000 (20K) features. The $\rm{AUC}_{10\%}$ value for $n_{\rm max}=0$, corresponding to classical performance evaluation, is also reported in the legend between square brackets.}
\label{wc-fig1}
\end{center}
\end{figure}

The results are reported in Fig.~\ref{wc-fig1}.
As expected, the $\rm{AUC}_{10\%}$ of each model decreases as $n_{\rm max}$ increases.
It drops to zero for $n_{\rm max}$ values between 30 and 50 (depending on the classifier): this means that all testing spam emails got misclassified as legitimate, after adding or obfuscating from 30 to 50 words.

The SVM and LR classifiers perform very similarly when they are not under attack (i.e., for $n_{\rm max} = 0$), regardless of the feature set size;
therefore, according to the viewpoint of classical performance evaluation, the designer could choose any of the eight models.
However, security evaluation highlights that they exhibit a very different robustness to the considered attack, since their $\rm{AUC}_{10\%}$ value decreases at very different rates as $n_{\rm max}$ increases; in particular, the LR classifier with 20,000 features clearly outperforms all the other ones, for all $n_{\rm max}$ values.
This result suggests the designer a very different choice than the one coming from classical performance evaluation:
the LR classifier with 20,000 features should be selected, given that it exhibit the same accuracy as the other ones in the absence of attacks, and a higher security under the considered attack.

\subsection{Biometric authentication}
\label{sect:exp-bio}

Multimodal biometric systems for personal identity recognition have received great interest in the past few years. It has been shown that combining information coming from different biometric traits can overcome the limits and the weaknesses inherent in every individual biometric, resulting in a higher accuracy.
Moreover, it is commonly believed that multimodal systems also improve security against spoofing attacks, which consist of claiming a false identity and submitting at least one fake biometric trait to the system (e.g., a ``gummy'' fingerprint or a photograph of a user's face).
The reason is that, to evade a multimodal system, one expects that the adversary should spoof \emph{all} the corresponding biometric traits. In this application example, we show how the designer of a multimodal system can verify if this hypothesis holds, before deploying the system, by simulating spoofing attacks against each of the matchers. To this end, we partially exploit the analysis in \cite{rodrigues09,johnson10}.

We consider a typical multimodal system, made up of a fingerprint and a face matcher, which operates as follows.
The design phase includes the enrollment of authorized users (clients): reference templates of their biometric traits are stored into a database, together with the corresponding identities.
During operation, each user provides the requested biometric traits to the sensors, and claims the identity of a client.
Then, each matcher compares the submitted trait with the template of the claimed identity, and provides a real-valued matching score: the higher the score, the higher the similarity.
We denote the score of the fingerprint and the face matcher respectively as $x_{\rm fing}$ and $x_{\rm face}$. Finally, the matching scores are combined through a proper fusion rule to decide whether the claimed identity is the user's identity (genuine user) or not (impostor).

As in \cite{rodrigues09,johnson10}, we consider the widely used likelihood ratio (LLR) score fusion rule \cite{nandakumar08}, and the matching scores as its features. Denoting as $\mathbf x$ the feature vector $(x_{\rm fing},x_{\rm face})$, the LLR can be written as:
\begin{equation}
\label{eq:LLR}
LLR(\mathbf x) = \\
\begin{cases}
{\rm L} & \text{if $p(\mathbf x | Y={\rm L}) / p(\mathbf x | Y={\rm M}) \geq t $,}
\\
{\rm M} &\text{if $p(\mathbf x | Y={\rm L}) / p(\mathbf x | Y={\rm M}) < t$,}
\end{cases}
\end{equation}
where L and M denote respectively the ``genuine'' and ``impostor'' class, and $t$ is a decision threshold set according to application requirements.
The distributions $p(x_{\rm fing}, x_{\rm face} | Y)$ are usually estimated from training data, while $t$ is estimated from a validation set.

\textbf{1) Attack scenario}. \textit{Goal}. In this case, each malicious user (impostor) aims at being accepted as a legitimate (genuine) one. This corresponds to a targeted integrity violation, where the adversary's goal is to maximize the matching score.

\textit{Knowledge}. As in \cite{rodrigues09,johnson10}, we assume that each impostor knows:
(k.i) the identity of the targeted client;
and (k.ii) the biometric traits used by the system.
No knowledge of (k.iii) the decision function and (k.iv) its parameters is assumed, and (k.v) no feedback is available from the classifier.

\textit{Capability}. We assume that: (c.i) spoofing attacks affect only testing data (exploratory attack); (c.ii) they do not affect the class priors;\footnote{Further, since the LLR considers only the ratio between the class-conditional distributions, changing the class priors is irrelevant.}
and, according to (k.i),
(c.iii) each adversary (impostor) controls her testing malicious samples.
We limit the capability of each impostor by assuming that (c.iv) only one specific trait can be spoofed at a time (i.e., only one feature can be manipulated), and that it is the same for all impostors.
We will thus repeat the evaluation twice, considering fingerprint and face spoofing, separately.
Defining how impostors can manipulate the features (in this case, the matching scores) in a spoofing attack is not straightforward.
The standard way of evaluating the security of biometric systems against spoofing attacks consists of fabricating fake biometric traits and assessing performance when they are submitted to the system.
However, this is a cumbersome task.
Here we follow the approach of \cite{rodrigues09,johnson10}, in which the impostor is assumed to fabricate ``perfect'' fake traits, i.e., fakes that produce the same matching score of the ``live'' trait of the targeted client.
This allows one to use the available genuine scores to simulate the fake scores, avoiding the fabrication of fake traits.
On the other hand, this assumption may be too pessimistic.
This limitation could be overcome by developing more realistic models of the distribution of the fake traits; however, this is a challenging research issue on its own, that is out of the scope of this work, and part of the authors' ongoing work \cite{biggio11-ijcb,biggio12-iet}.

\textit{Attack strategy}. The above attack strategy modifies only the testing data, and: (a.i) it does not modify the class priors; (a.ii) it does not affect the genuine class, but all impostors; and (a.iii) any impostor spoofs the considered biometric trait (either the face or fingerprint) of the known identity.
According to the above assumption, as in \cite{rodrigues09,johnson10}, any spoofing attack is simulated by replacing the corresponding impostor score (either the face or fingerprint) with the score of the targeted genuine user (chosen at random from the legitimate samples in the testing set).
This can be represented with a function $\mathcal A(\mathbf x)$, that, given an impostor $\mathbf x = (x_{\rm fing}, x_{\rm face})$, returns either a vector $(x^{\prime}_{\rm fing}, x_{\rm face})$ (for fingerprint spoofing) or $(x_{\rm fing}, x^{\prime}_{\rm face})$ (for face spoofing), where $x^{\prime}_{\rm fing}$ and $x^{\prime}_{\rm face}$ are the matching scores of the targeted genuine user.

\textbf{2) Data model}. Since training data is not affected, we set $p_{\rm tr}(\mathbf X,Y) = p_{\mathcal D}(\mathbf X,Y)$.
Then, we set (a.i) $p_{\rm ts}(Y)=p_{\mathcal D}(Y)$;
and (a.ii) $p_{\rm ts}(A={\rm T}|Y={\rm L})=0$ and $p_{\rm ts}(A={\rm T}|Y={\rm M})=1$.
Thus, $p_{\rm ts}(\mathbf X | Y={\rm L}) = p_{\mathcal D}(\mathbf X | Y={\rm L})$, while all the malicious samples in TS have to be manipulated by $\mathcal A(\mathbf x)$, according to (a.iii). This amounts to \emph{empirically} defining $p_{\rm ts}(\mathbf X | Y={\rm M}, A={\rm T})$. As in the spam filtering case, $p_{\rm ts}(\mathbf X | Y=y, A={\rm F})$, for $y={\rm L,M}$, will be empirically approximated from the available data in $\mathcal D$.

The definition of the above attack scenario and data model is summarized in Table~\ref{table:exp} (second column).

\textbf{3) Construction of TR and TS}.
We use the NIST Biometric Score Set, Release 1.\footnote{\url{http://www.itl.nist.gov/iad/894.03/biometricscores/}}
It contains raw similarity scores obtained on a set of 517 users from two different face matchers (named `G' and `C'), and from one fingerprint matcher using the left and right index finger. For each user, one genuine score and 516 impostor scores are available for each matcher and each modality, for a total of 517 genuine and 266,772 impostor samples.
We use the scores of the `G' face matcher and the ones of the fingerprint matcher for the left index finger, and normalize them in $[0,1]$ using the min-max technique.
The data set $\mathcal D$ is made up of pairs of face and fingerprint scores, each belonging to the same user.
We first randomly subdivide $\mathcal D$ into a disjoint training and testing set, $\mathcal D_{\rm TR}$ and $\mathcal D_{\rm TS}$, containing respectively 80\% and 20\% of the samples.
As the attack does not affect the training samples, we set TR$=\mathcal D_{\rm TR}$.
The sets $\mathcal D_{\rm TS}^{\rm L,F}$ and $\mathcal D_{\rm TS}^{\rm M,F}$ are constructed using $\mathcal D_{\rm TS}$, while $\mathcal D_{\rm TS}^{\rm L,F} = \emptyset$.
The set of attack samples $\mathcal D_{\rm TS}^{\rm M,T}$ is obtained by modifying each sample of $\mathcal D_{\rm TS}^{\rm M,F}$ with $\mathcal A(\mathbf x)$.
We finally set the size of TS as $| \mathcal D_{\rm TS}|$, and run Algorithm \ref{alg:data-set-construction} to obtain it.

\textbf{4) Performance evaluation.}
In biometric authentication tasks, the performance is usually measured in terms of genuine acceptance rate (GAR) and false acceptance rate (FAR), respectively the fraction of genuine and impostor attempts that are accepted as genuine by the system. We use here the complete ROC curve, which shows the GAR as a function of the FAR for all values of the decision threshold $t$ (see Eq.~\ref{eq:LLR}).

To estimate $p(\mathbf X | Y=y)$, for $y={\rm L,M}$, in the LLR score fusion rule (Eq.~\ref{eq:LLR}), we assume that $x_{\rm fing}$ and $x_{\rm face}$ are conditionally independent given $Y$, as usually done in biometrics. We thus compute a maximum likelihood estimate of $p(x_{\rm fing},x_{\rm face} | Y)$ from TR using a product of two Gamma distributions, as in \cite{rodrigues09}.

\begin{figure}[t]
\begin{center}
\includegraphics[width=0.4\textwidth]{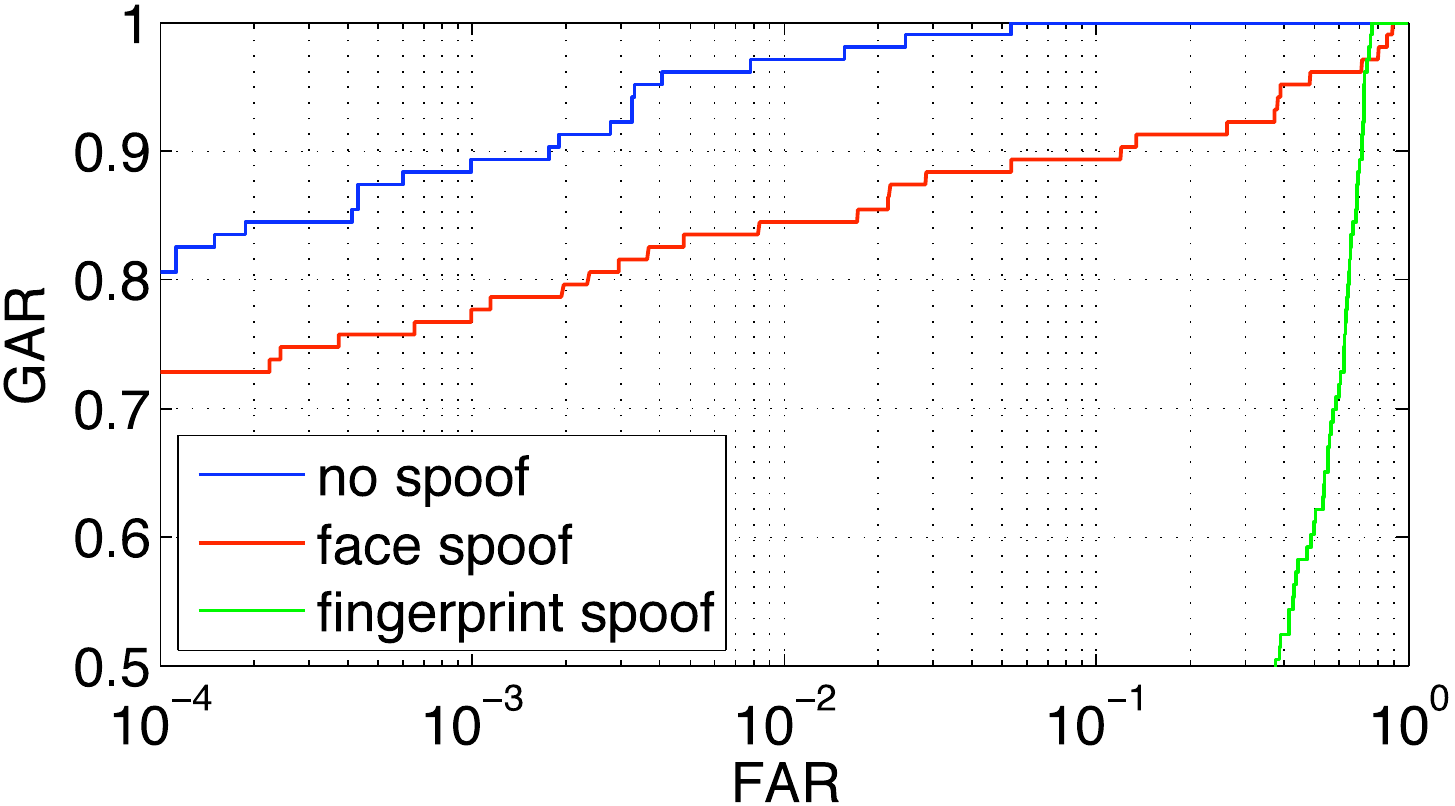}
\caption{ROC curves of the considered multimodal biometric system, under a simulated spoof attack against the fingerprint or the face matcher.}
\label{fig:exp-bio}
\end{center}
\end{figure}

Fig.~\ref{fig:exp-bio} shows the ROC curve evaluated with the standard approach (without spoofing attacks), and the ones corresponding to the simulated spoofing attacks. The FAR axis is in logarithmic scale to focus on low FAR values, which are the most relevant ones in security applications.
For any operational point on the ROC curve (namely, for any value of the decision threshold $t$), the effect of the considered spoofing attack is to increase the FAR, while the GAR does not change. This corresponds to a shift of the ROC curve to the right. Accordingly, the FAR under attack must be compared with the original one, for the same GAR.

Fig.~\ref{fig:exp-bio} clearly shows that the FAR of the biometric system significantly increases under the considered attack scenario, especially for fingerprint spoofing.
As far as this attack scenario is deemed to be realistic, the system designer should conclude that the considered system can be evaded by spoofing only one biometric trait, and thus does not exhibit a higher security than each of its individual classifiers.
For instance, in applications requiring relatively high security, a reasonable choice may be to chose the operational point with FAR=$10^{-3}$ and GAR=0.90, using classical performance evaluation, i.e., without taking into account spoofing attacks.
This means that the probability that the deployed system wrongly accepts an impostor as a genuine user (the FAR) is expected to be 0.001.
However, under face spoofing, the corresponding FAR increases to $0.10$, while it jumps to about $0.70$ under fingerprint spoofing.
In other words, an impostor who submits a perfect replica of the face of a client has a probability of 10\% of being accepted as genuine, while this probability is as high as 70\%, if she is able to perfectly replicate the fingerprint.
This is unacceptable for security applications, and provides further support to the conclusions of \cite{rodrigues09,johnson10} against the common belief that multimodal biometric systems are intrinsically more robust than unimodal systems.

\subsection{Network intrusion detection}
\label{sect:exp-ids}

Intrusion detection systems (IDSs) analyze network traffic to prevent and detect malicious activities like intrusion attempts, port scans, and denial-of-service attacks.\footnote{The term ``attack'' is used in this field to denote a malicious activity, even when there is no deliberate attempt of misleading an IDS. In adversarial classification, as in this paper, this term is used to specifically denote the attempt of misleading a classifier, instead. To avoid any confusion, in the following we will refrain from using the term ``attack'' with the former meaning, using paraphrases, instead.}
When suspected malicious traffic is detected, an alarm is raised by the IDS and subsequently handled by the system administrator.
Two main kinds of IDSs exist: misuse detectors and anomaly-based ones.
\emph{Misuse detectors} match the analyzed network traffic against a database of signatures of known malicious activities (e.g., \verb=Snort=).\footnote{\url{http://www.snort.org/}}
The main drawback is that they are not able to detect never-before-seen malicious activities, or even variants of known ones.
To overcome this issue, \emph{anomaly-based} detectors have been proposed.
They build a statistical model of the normal traffic using machine learning techniques, usually one-class classifiers (e.g., \verb=PAYL=~\cite{wang04}), and raise an alarm when anomalous traffic is detected.
Their training set is constructed, and periodically updated to follow the changes of normal traffic, by collecting unsupervised network traffic during operation, assuming that it is normal (it can be filtered by a misuse detector, and should be discarded if some system malfunctioning occurs during its collection).
This kind of IDS is vulnerable to causative attacks, since an attacker may inject carefully designed malicious traffic during the collection of training samples to force the IDS to learn a wrong model of the normal traffic \cite{barreno-ASIACCS06,laskov09,kloft10,rubinstein09,biggio12-icml}.

Here we assume that an anomaly-based IDS is being designed, using a one-class $\nu$-SVM classifier with a radial basis function (RBF) kernel and the feature vector representation proposed in~\cite{wang04}.
Each network packet is considered as an individual sample to be labeled as normal (legitimate) or anomalous (malicious), and is represented as a 256-dimensional feature vector, defined as the histogram of byte frequencies in its payload (this is known as ``1-gram'' representation in the IDS literature).
We then focus on the model selection stage. In the above setting, it amounts to choosing the values of the $\nu$ parameter of the learning algorithm (which is an upper bound on the false positive error rate on training data \cite{scholkopf00}), and the $\gamma$ value of the RBF kernel.
For the sake of simplicity, we assume that $\nu$ is set to $0.01$ as suggested in~\cite{perdisci-ICDM06}, so that only $\gamma$ has to be chosen.

We show how the IDS designer can select a model (the value of $\gamma$) based also on the evaluation of classifier security.
We focus on a causative attack similar to the ones considered in \cite{cardenas06}, aimed at forcing the learned model of normal traffic to include samples of intrusions to be attempted during operation.
To this end, the attack samples should be carefully designed such that they include some features of the desired intrusive traffic, but do not perform any real intrusion (otherwise, the collected traffic may be discarded, as explained above).

\textbf{1) Attack scenario}. \textit{Goal}. 
This attack aims to cause an indiscriminate integrity violation by maximizing the fraction of malicious testing samples misclassified as legitimate.

\textit{Knowledge}. The adversary is assumed to know: (k.ii) the feature set; 
and (k.iii) that a one-class classifier is used. No knowledge of (k.i) the training data and (k.iv) the classifiers' parameters is available to the adversary, as well as (k.v) any feedback from the classifier.

\textit{Capability}.
The attack consists of injecting \emph{malicious} samples into the training set.
Accordingly, we assume that: (c.i) the adversary can inject malicious samples into the training data, without manipulating testing data (causative attack);
(c.ii) she can modify the class priors by injecting a maximum fraction $p_{\rm max}$ of malicious samples into the training data;
(c.iii) all the injected malicious samples can be manipulated;
and (c.iv) the adversary is able to completely control the feature values of the malicious attack samples.
As in \cite{laskov09,kloft10,nelson08,rubinstein09}, we repeat the security evaluation for $p_{\rm max} \in [0, 0.5]$, since it is unrealistic that the adversary can control the majority of the training data.

\textit{Attack strategy}.
The goal of maximizing the percentage of malicious testing samples misclassified as legitimate, for any $p_{\rm max}$ value, can be attained by manipulating the training data with the following attack strategy:
(a.i) the maximum percentage of malicious samples $p_{\rm max}$ will be injected into the training data;
(a.ii) all the injected malicious samples will be attack samples, while no legitimate samples will be affected;
and (a.iii) the feature values of the attack samples will be equal to that of the malicious testing samples.
To understand the latter assumption, note that: first, the best case for the adversary is when the attack samples exhibit the same histogram of the payload's byte values as the malicious testing samples, as this intuitively forces the model of the normal traffic to be as similar as possible to the distribution of the testing malicious samples; and, second, this does not imply that the attack samples perform any intrusion, allowing them to avoid detection.
In fact, one may construct an attack packet which exhibit the same feature values of an intrusive network packet (i.e., a malicious testing sample), but looses its intrusive functionality, by simply shuffling its payload bytes.
Accordingly, for our purposes, we directly set the feature values of the attack samples equal to those of the malicious testing samples, without constructing real network packets.

\textbf{2) Data model}. Since testing data is not affected by the considered causative attack, we set $p_{\rm ts}(\mathbf X, Y) = p_{\mathcal D}(\mathbf X,Y)$.
We then encode the attack strategy as follows.
According to (a.ii), $p_{\rm tr}(A={\rm T} | Y={\rm L})=0$ and $p_{\rm tr}(A={\rm T} | Y={\rm M})=1$.
The former assumption implies that $p_{\rm tr}(\mathbf X | Y={\rm L}) = p_{\mathcal D}(\mathbf X | Y={\rm L})$.
Since the training data of one-class classifiers does not contain any malicious samples (see explanation at the beginning of this section), i.e., $p_{\rm tr}(A={\rm F}|Y={\rm M})=0$, the latter assumption implies that the class prior $p_{\rm tr}(Y={\rm M})$ corresponds exactly to the fraction of attack samples in the training set.
Therefore, assumption (a.i) amounts to setting $p_{\rm tr}(Y={\rm M})=p_{\rm max}$.
Finally, according to (a.iii), the distribution of attack samples $p_{\rm tr}(\mathbf X | Y={\rm M},A={\rm T})$ will be defined as the \emph{empirical} distribution of the malicious testing samples, namely,  $\mathcal D_{\rm TR}^{{\rm M},{\rm T}} = \mathcal D_{\rm TS}^{{\rm M},{\rm F}}$.
The distribution $p_{\rm tr}(\mathbf X | Y={\rm L},A={\rm F})=p_{\mathcal D}(\mathbf X | {\rm L})$ will also be empirically defined as the set $\mathcal D_{\rm TR}^{{\rm L},{\rm F}}$.

The definition of the attack scenario and data model is summarized in Table~\ref{table:exp} (third column).

\textbf{3) Construction of TR and TS}.
We use the data set of \cite{perdisci-ICDM06}. It consists of 1,699,822 legitimate samples (network packets) collected by a web server during five days in 2006, and a publicly available set of 205 malicious samples coming from intrusions which exploit the HTTP protocol \cite{ingham07}.\footnote{\url{http://www.i-pi.com/HTTP-attacks-JoCN-2006/}}
To construct TR and TS, we take into account the chronological order of network packets as in Sect.~\ref{sect:exp-spam}, and the fact that in this application all the malicious samples in $\mathcal D$ must be inserted in the testing set \cite{ingham07,cretu08}.
Accordingly, we set $\mathcal D_{\rm TR}$ as the first 20,000 legitimate packets of day one, and $\mathcal D_{\rm TS}$ as the first 20,000 legitimate samples of day two, plus all the malicious samples.
Since this attack does not affect testing samples, we set TS $=\mathcal D_{\rm TS}$.
We then set $\mathcal D_{\rm TR}^{\rm L,F} = \mathcal D_{\rm TR}$, and $\mathcal D_{\rm TR}^{\rm M,F} = \mathcal D_{\rm TR}^{\rm L,T} = \emptyset$. 
Since the feature vectors of attack samples are identical to those of the malicious samples, as above mentioned, we set $\mathcal D_{\rm TR}^{\rm M,T} = \mathcal D_{\rm TS}^{\rm M,F}$ (where the latter set clearly includes the malicious samples in $\mathcal D_{\rm TS}$). 
The size of TR is initially set to 20,000. We then consider different attack scenarios by increasing the number of attack samples in the training data, up to 40,000 samples in total, that corresponds to $p_{\rm max}=0.5$.
For each value of $p_{\rm max}$, TR is obtained by running Algorithm \ref{alg:data-set-construction} with the proper inputs.

\textbf{4) Performance evaluation}.
Classifier performance is assessed using the ${\rm AUC}_{10\%}$ measure, for the same reasons as in Sect.~\ref{sect:exp-spam}.
The performance under attack is evaluated as a function of $p_{\rm max}$, as in \cite{laskov09,nelson08}, which reduces to the classical performance evaluation when $p_{\rm max}=0$.
For the sake of simplicity, we consider only two values of the parameter $\gamma$, which clearly point out how design choices based only on classical performance evaluation methods can be unsuitable for adversarial environments.

\begin{figure}[t]
\begin{center}
\includegraphics[width=0.42\textwidth]{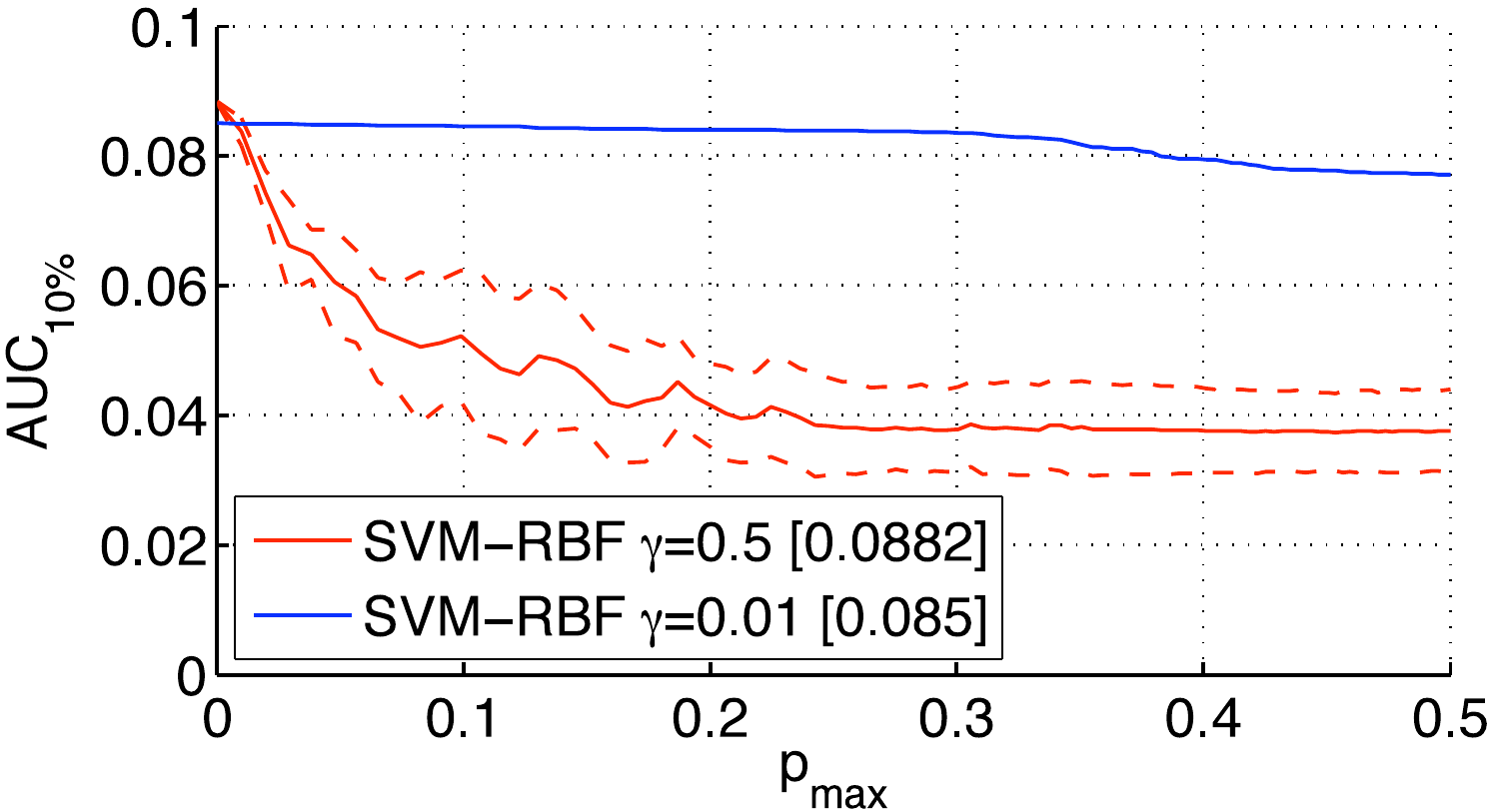}
\caption{$\rm{AUC}_{10\%}$ as a function of $p_{\rm max}$, for the $\nu$-SVMs with RBF kernel. The $\rm{AUC}_{10\%}$ value for $p_{\rm max}=0$, corresponding to classical performance evaluation, is also reported in the legend between square brackets. The standard deviation (dashed lines) is reported only for $\gamma=0.5$, since it was negligible for $\gamma=0.01$.}
\label{fig:exp-ids}
\end{center}
\end{figure}

The results are reported in Fig.~\ref{fig:exp-ids}.
In the absence of attacks ($p_{\rm max}=0$), the choice $\gamma=0.5$ appears slightly better than $\gamma=0.01$.
Under attack, the performance for $\gamma=0.01$ remains almost identical as the one without attack, and starts decreasing very slightly only when the percentage of attack samples in the training set exceeds 30\%.
On the contrary, for $\gamma=0.5$ the performance suddenly drops as $p_{\rm max}$ increases, becoming lower than the one for $\gamma=0.01$ when $p_{\rm max}$ is as small as about 1\%.
The reason for this behavior is the following. The attack samples can be considered outliers with respect to the legitimate training samples, and, for large $\gamma$ values of the RBF kernel, the SVM discriminant function tends to overfit, forming a ``peak'' around each individual training sample. Thus, it exhibits relatively high values also in the region of the feature space where the attack samples lie, and this allows many of the corresponding testing intrusions.
Conversely, this is not true for lower $\gamma$ values, where the higher spread of the RBF kernel leads to a smoother discriminant function, which exhibits much lower values for the attack samples.

According to the above results, the choice of $\gamma = 0.5$, suggested by classical performance evaluation for $p{\rm max}=0$ is clearly unsuitable from the viewpoint of classifier security under the considered attack, unless it is deemed unrealistic that an attacker can inject more than 1\% attack samples into the training set.
To summarize, the designer should select $\gamma=0.01$ and trade a small decrease of classification accuracy in the absence of attacks for a significant security improvement.

\section{Secure design cycle: next steps}
\label{sect:secure-design}

The classical design cycle of a pattern classifier \cite{duda-hart-stork} consists of: data collection, data pre-processing, feature extraction and selection, model selection (including the choice of the learning and classification algorithms, and the tuning of their parameters), and performance evaluation.
We pointed out that this design cycle disregards the threats that may arise in adversarial settings, and extended the performance evaluation step to such settings. Revising the remaining steps under a security viewpoint remains a very interesting issue for future work.
Here we briefly outline how this open issue can be addressed.

If the adversary is assumed to have some control over the data collected for classifier training and parameter tuning, a filtering step to detect and remove attack samples should also be performed (see, e.g., the \emph{data sanitization} method of \cite{cretu08}).
 
Feature extraction algorithms should be designed to be robust to sample manipulation. Alternatively, features which are more difficult to be manipulated should be used.
For instance, in \cite{sculley06} inexact string matching was proposed to counteract word obfuscation attacks in spam filtering. In biometric recognition, it is very common to use additional input features to detect the presence (``liveness detection'') of attack samples coming from spoofing attacks, i.e., fake biometric traits \cite{encyclopedia-bio09}.
The adversary could also undermine feature selection, e.g., to force the choice of a set of features that are easier to manipulate, or that are not discriminant enough with respect to future attacks.
Therefore, feature selection algorithms should be designed by taking into account not only the discriminant capability, but also the robustness of features to adversarial manipulation.

Model selection is clearly the design step that is more subject to attacks. Selected algorithms should be robust to causative and exploratory attacks.
In particular, robust learning algorithms should be adopted, if no data sanitization can be performed.
The use of robust statistics has already been proposed to this aim; in particular, to devise learning algorithms robust to a limited amount of data contamination \cite{rubinstein09,huang11}, and classification algorithms robust to specific exploratory attacks \cite{globerson-ICML06,jorgensen08,kolcz09,biggio11-smc}.

Finally, a secure system should also guarantee the \emph{privacy} of its users, against attacks aimed at stealing confidential information \cite{huang11}.
For instance, privacy preserving methods have been proposed in biometric recognition systems to protect the users against the so-called hill-climbing attacks, whose goal is to get information about the users' biometric traits \cite{adler05,encyclopedia-bio09}. Randomization of some classifierÕs parameters has been also proposed to preserve privacy in \cite{huang11,biggio08-spr}.

\section{Contributions, limitations \\ and open issues}
\label{sect:open-issues}

In this paper we focused on empirical security evaluation of pattern classifiers that have to be deployed in adversarial environments, and proposed how to revise the classical performance evaluation design step, which is not suitable for this purpose.

Our main contribution is a framework for empirical security evaluation that formalizes and generalizes ideas from previous work, and can be applied to different classifiers, learning algorithms, and classification tasks. It is grounded on a formal
model of the adversary, and on a model of data distribution that can represent all the attacks considered in previous work; provides a systematic method for the generation of training and testing sets that enables security evaluation;
and can accommodate application-specific techniques for attack simulation. This is a clear advancement with respect to previous work, since without a general framework most of the proposed techniques (often tailored to a given classifier model, attack, and application) could not be directly applied to other problems.

An intrinsic limitation of our work is that security evaluation is carried out empirically, and it is thus data-dependent; on the other hand, model-driven analyses \cite{cardenas-ws06,cardenas06,laskov09} require a full analytical model of the problem and of the adversary's behavior, that may be very difficult to develop for real-world applications.
Another intrinsic limitation is due to fact that our method is not application-specific, and, therefore, provides only high-level guidelines for simulating attacks. Indeed, detailed guidelines require one to take into account application-specific constraints and adversary models. Our future work will be devoted to develop techniques for simulating attacks for different applications. 

Although the design of secure classifiers is a distinct problem than security evaluation, our framework could be also exploited to this end. For instance, simulated attack samples can be included into the training data to improve security of \emph{discriminative} classifiers (e.g., SVMs), while the proposed data model can be exploited to design more secure \emph{generative} classifiers. We obtained encouraging preliminary results on this topic \cite{biggio11-smc}.

\section*{Acknowledgments}
The authors are grateful to Davide Ariu, Gavin Brown, Pavel Laskov, and Blaine Nelson for discussions and comments on an earlier version of this paper.
This work was partly supported by a grant awarded to Battista Biggio by Regione Autonoma della Sardegna, PO Sardegna FSE 2007-2013, L.R.~7/2007 ``Promotion of the scientific research and technological innovation in Sardinia'', by the project CRP-18293 funded by Regione Autonoma della Sardegna, L.R. 7/2007, Bando 2009, and by the TABULA RASA project, funded within the 7th Framework Research Programme of the European Union.


\vspace{-1.1 cm}

\begin{biography}
[{\includegraphics[width=1in,height =1.25in,clip,keepaspectratio]{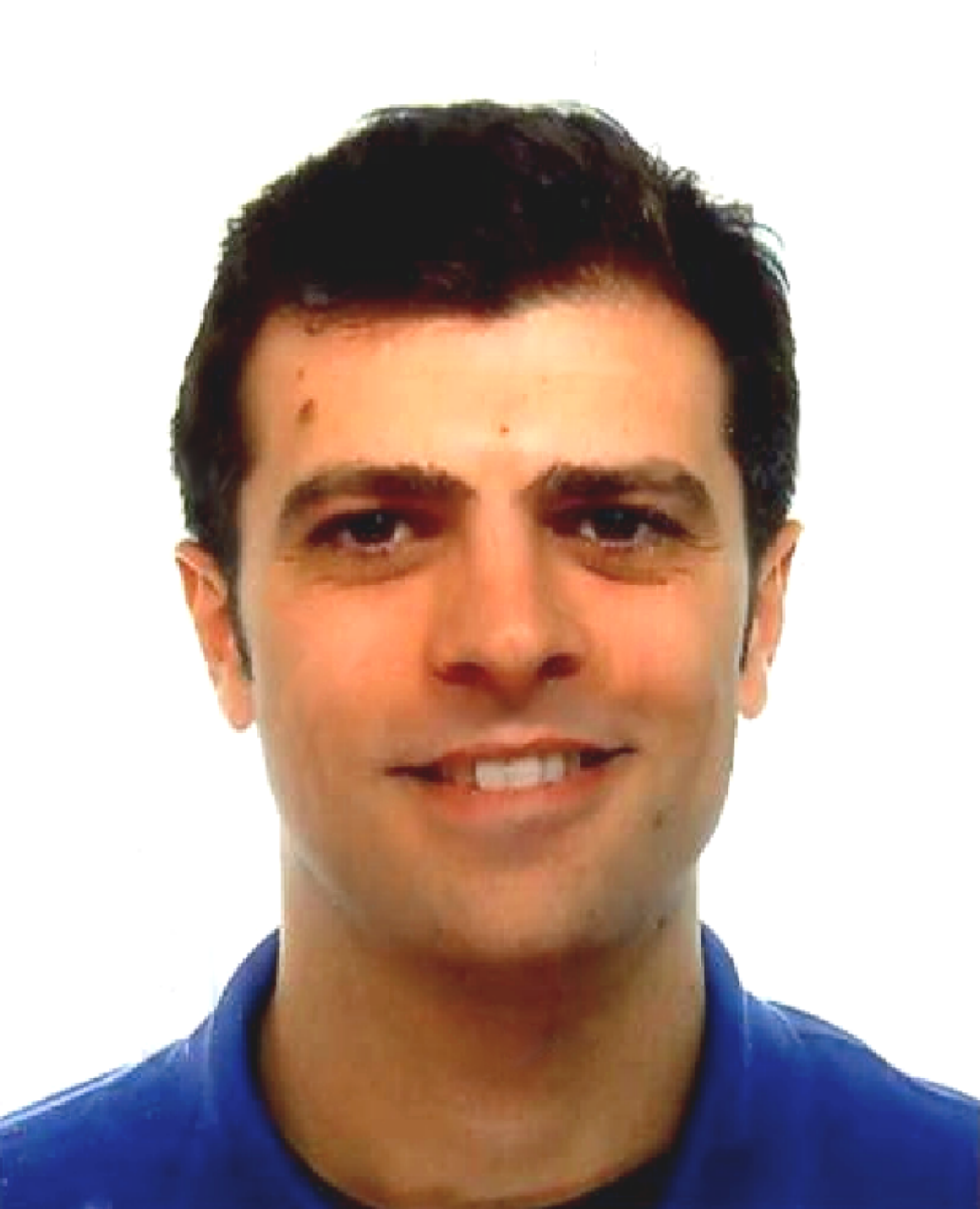}}]{Battista Biggio} received the M. Sc. degree in Electronic Eng., with honors, and the Ph. D. in Electronic Eng. and Computer Science, respectively in 2006 and 2010, from the University of Cagliari, Italy.
Since 2007 he has been working for the Dept. of Electrical and Electronic Eng. of the same University, where he holds now a postdoctoral position. 
From May 12th, 2011 to November 12th, 2011, he visited the University of T\"ubingen, Germany, and worked on the security of machine learning algorithms to contamination of training data.
His research interests currently include: secure / robust machine learning and pattern recognition methods, multiple classifier systems, kernel methods, biometric authentication, spam filtering, and computer security.
He serves as a reviewer for several international conferences and journals, including Pattern Recognition and Pattern Recognition Letters.
Dr. Biggio is a member of the IEEE Institute of Electrical and Electronics Engineers (Computer Society), and of the Italian Group of Italian Researchers in Pattern Recognition (GIRPR), affiliated to the International Association for Pattern Recognition.
\end{biography}

\vspace{-1.2 cm}

\begin{biography}[{\includegraphics[width=1in,height =1.25in,clip,keepaspectratio]{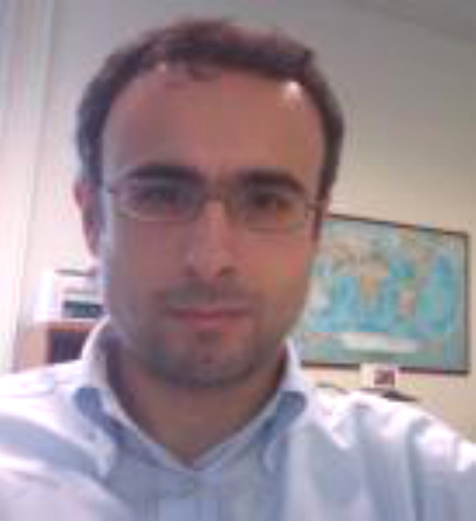}}]{Giorgio Fumera} received the M. Sc. degree in Electronic Eng., with honors, and the Ph.D. degree in Electronic Eng. and Computer Science, respectively in 1997 and 2002, from the University of Cagliari, Italy. Since February 2010 he is Associate Professor of Computer Eng. at the Dept. of Electrical and Electronic Eng. of the same University.
His research interests are related to methodologies and applications of statistical pattern recognition, and include multiple classifier systems, classification with the reject option, adversarial classification and document categorization. On these topics he published more than sixty papers in international journal and conferences. He acts as reviewer for the main international journals in this field, including IEEE Trans. on Pattern Analysis and Machine Intelligence, Journal of Machine Learning Research, and Pattern Recognition.
Dr. Fumera is a member of the IEEE Institute of Electrical and Electronics Engineers (Computer Society), of the Italian Association for Artificial Intelligence (AI*IA), and of the Italian Group of Italian Researchers in Pattern Recognition (GIRPR), affiliated to the International Association for Pattern Recognition.
\end{biography}

\vspace{-1.2 cm}

\begin{biography}[{\includegraphics[width=1in,height =1.25in,clip,keepaspectratio]{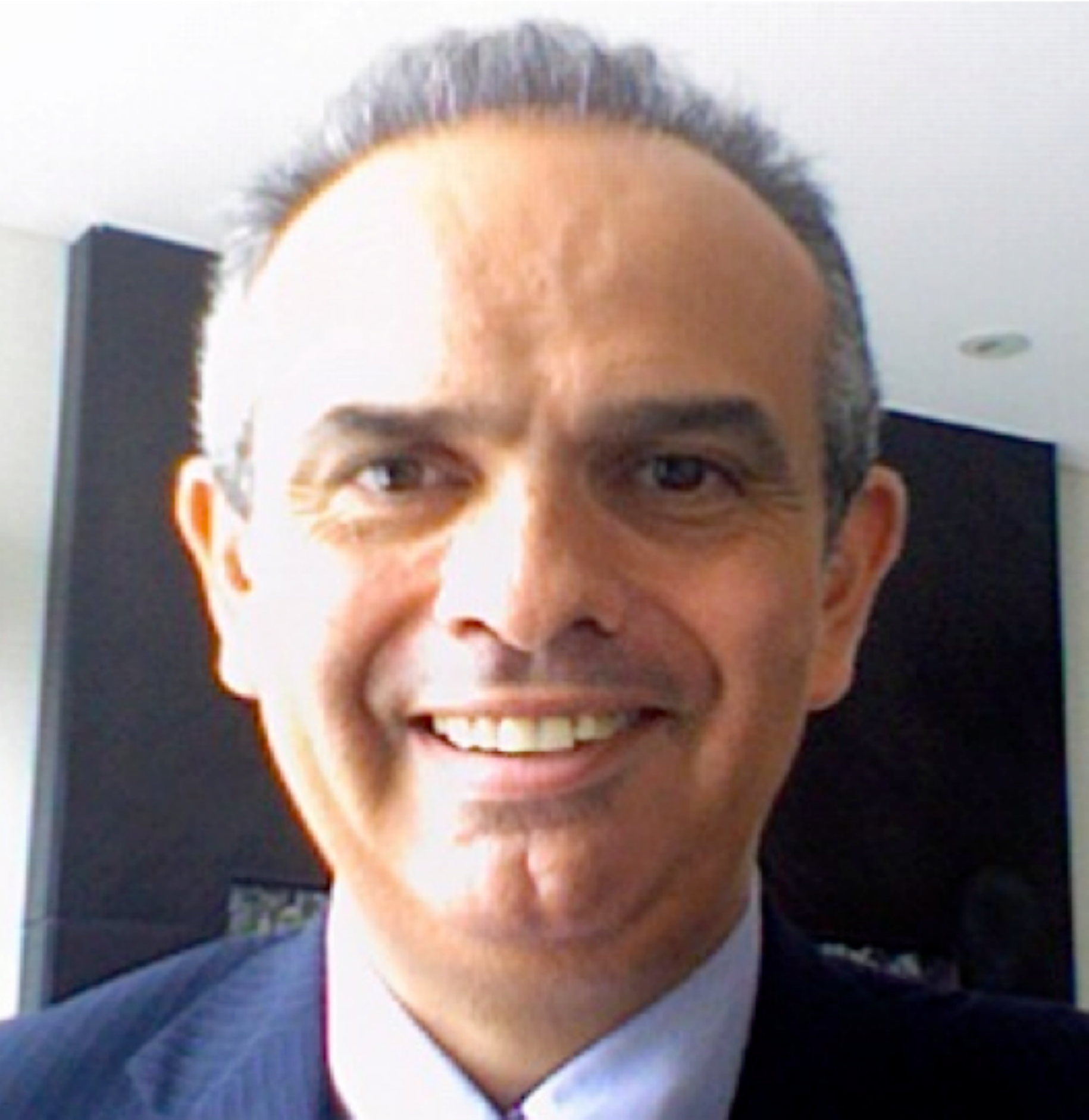}}]{Fabio Roli} received his M. Sc. degree, with honors, and Ph. D. degree in Electronic Eng. from the University of Genoa, Italy. He was a member of the research group on Image Processing and Understanding of the University of Genoa, Italy, from 1988 to 1994. He was adjunct professor at the University of Trento, Italy, in 1993 and 1994. In 1995, he joined the Dept. of Electrical and Electronic Eng. of the University of Cagliari, Italy, where he is now professor of computer engineering and head of the research group on pattern recognition and applications. His research activity is focused on the design of pattern recognition systems and their applications to biometric personal identification, multimedia text categorization, and computer security. On these topics, he has published more than two hundred papers at conferences and on journals. He was a very active organizer of international conferences and workshops, and established the popular workshop series on multiple classifier systems. Dr. Roli is a member of the governing boards of the International Association for Pattern Recognition and of the IEEE Systems, Man and Cybernetics Society. He is Fellow of the IEEE, and Fellow of the International Association for Pattern Recognition.
\end{biography}


\begin{thebibliography}{10}
\providecommand{\url}[1]{#1}
\csname url@samestyle\endcsname
\providecommand{\newblock}{\relax}
\providecommand{\bibinfo}[2]{#2}
\providecommand{\BIBentrySTDinterwordspacing}{\spaceskip=0pt\relax}
\providecommand{\BIBentryALTinterwordstretchfactor}{4}
\providecommand{\BIBentryALTinterwordspacing}{\spaceskip=\fontdimen2\font plus
\BIBentryALTinterwordstretchfactor\fontdimen3\font minus
  \fontdimen4\font\relax}
\providecommand{\BIBforeignlanguage}[2]{{%
\expandafter\ifx\csname l@#1\endcsname\relax
\typeout{** WARNING: IEEEtran.bst: No hyphenation pattern has been}%
\typeout{** loaded for the language `#1'. Using the pattern for}%
\typeout{** the default language instead.}%
\else
\language=\csname l@#1\endcsname
\fi
#2}}
\providecommand{\BIBdecl}{\relax}
\BIBdecl

\bibitem{rodrigues09}
R.~N. Rodrigues, L.~L. Ling, and V.~Govindaraju, ``Robustness of multimodal
  biometric fusion methods against spoof attacks,'' \emph{J. Vis. Lang.
  Comput.}, vol.~20, no.~3, pp. 169--179, 2009.

\bibitem{johnson10}
P.~Johnson, B.~Tan, and S.~Schuckers, ``Multimodal fusion vulnerability to
  non-zero effort (spoof) imposters,'' in \emph{IEEE Int'l Workshop on Inf. Forensics and
  Security}, 2010, pp. 1--5.

\bibitem{fogla06}
P.~Fogla, M.~Sharif, R.~Perdisci, O.~Kolesnikov, and W.~Lee, ``Polymorphic
  blending attacks,'' in \emph{Proc. 15th Conf. on USENIX Security Symp.}\hskip 1em plus 0.5em minus 0.4em\relax
  CA, USA: USENIX Association, 2006.

\bibitem{wittel04}
G.~L. Wittel and S.~F. Wu, ``On attacking statistical spam filters,'' in
  \emph{1st Conf. on Email and Anti-Spam}, CA, USA, 2004.

\bibitem{lowd05-ceas}
D.~Lowd and C.~Meek, ``Good word attacks on statistical spam filters,'' in
  \emph{2nd Conf. on Email and Anti-Spam}, CA, USA, 2005.

\bibitem{kolcz09}
A.~Kolcz and C.~H. Teo, ``Feature weighting for improved classifier
  robustness,'' in \emph{6th Conf. on Email and Anti-Spam}, CA, USA, 2009.

\bibitem{skillicorn09}
D.~B. Skillicorn, ``Adversarial knowledge discovery,'' \emph{IEEE Intell.
  Syst.}, vol.~24, pp. 54--61, 2009.

\bibitem{fetterly07}
D.~Fetterly, ``Adversarial information retrieval: The manipulation of web
  content,'' \emph{ACM Computing Reviews}, 2007.

\bibitem{duda-hart-stork}
R.~O. Duda, P.~E. Hart, and D.~G. Stork, \emph{Pattern Classification}.\hskip
  1em plus 0.5em minus 0.4em\relax Wiley-Interscience Publication, 2000.

\bibitem{dalvi04}
N.~Dalvi, P.~Domingos, Mausam, S.~Sanghai, and D.~Verma, ``Adversarial
  classification,'' in \emph{10th ACM SIGKDD Int'l Conf. on Knowl.
  Discovery and Data Mining}, WA, USA, 2004, pp. 99--108.

\bibitem{barreno-ASIACCS06}
M.~Barreno, B.~Nelson, R.~Sears, A.~D. Joseph, and J.~D. Tygar, ``Can machine
  learning be secure?'' in \emph{Proc. Symp. Inf., Computer and Commun. Sec. (ASIACCS)}.\hskip 1em
  plus 0.5em minus 0.4em\relax NY, USA: ACM, 2006, pp. 16--25.

\bibitem{cardenas-ws06}
A.~A. C\'{a}rdenas and J.~S. Baras, ``Evaluation of classifiers: Practical
  considerations for security applications,'' in \emph{AAAI Workshop on
  Evaluation Methods for Machine Learning}, MA, USA, 2006.

\bibitem{laskov10-ed}
P.~Laskov and R.~Lippmann, ``Machine learning in adversarial environments,''
  \emph{Machine Learning}, vol.~81, pp. 115--119, 2010.

\bibitem{huang11}
L.~Huang, A.~D. Joseph, B.~Nelson, B.~Rubinstein, and J.~D. Tygar,
  ``Adversarial machine learning,'' in \emph{4th ACM Workshop on Artificial
  Intelligence and Security}, IL, USA, 2011, pp. 43--57.

\bibitem{barreno10}
M.~Barreno, B.~Nelson, A.~Joseph, and J.~Tygar, ``The security of machine
  learning,'' \emph{Machine Learning}, vol.~81, pp. 121--148, 2010.

\bibitem{lowd05}
D.~Lowd and C.~Meek, ``Adversarial learning,'' in \emph{Proc. 11th 
  ACM SIGKDD Int'l Conf. on Knowl. Discovery and Data
  Mining}, A.~Press, Ed., IL, USA, 2005, pp. 641--647.

\bibitem{laskov09}
P.~Laskov and M.~Kloft, ``A framework for quantitative security analysis of
  machine learning,'' in \emph{Proc. 2nd ACM Workshop on
  Security and Artificial Intelligence}.\hskip 1em plus 0.5em minus
  0.4em\relax NY, USA: ACM, 2009, pp. 1--4.

\bibitem{nips07-adv}
\BIBentryALTinterwordspacing
P.~Laskov and R.~Lippmann, Eds., \emph{{NIPS} {W}orkshop on {M}achine
  {L}earning in {A}dversarial {E}nvironments for {C}omputer {S}ecurity}, 2007.
  [Online]. Available: \url{http://mls-nips07.first.fraunhofer.de/}
\BIBentrySTDinterwordspacing

\bibitem{dagstuhl12-adv}
\BIBentryALTinterwordspacing
A.~D. Joseph, P.~Laskov, F.~Roli, and D.~Tygar, Eds., \emph{{D}agstuhl
  {P}erspectives {W}orkshop on {M}ach. {L}earning {M}ethods for {C}omputer
  {S}ec.}, 2012. [Online]. Available: \url{http://www.dagstuhl.de/12371/}
\BIBentrySTDinterwordspacing

\bibitem{kuncheva07}
A.~M. Narasimhamurthy and L.~I. Kuncheva, ``A framework for generating data to
  simulate changing environments,'' in \emph{Artificial Intell. and
  Applications}.\hskip 1em plus 0.5em minus 0.4em\relax
  IASTED/ACTA Press, 2007, pp. 415--420.

\bibitem{rizzi09}
S.~Rizzi, ``What-if analysis,'' \emph{Enc. of Database Systems}, pp.
  3525--3529, 2009.

\bibitem{newsome06}
J.~Newsome, B.~Karp, and D.~Song, ``Paragraph: Thwarting signature learning by
  training maliciously,'' in \emph{Recent Advances in Intrusion Detection},
  ser. LNCS.\hskip 1em plus 0.5em minus 0.4em\relax Springer, 2006, pp.
  81--105.

\bibitem{globerson-ICML06}
A.~Globerson and S.~T. Roweis, ``Nightmare at test time: robust learning by
  feature deletion,'' in \emph{Proc. 23rd Int'l Conf. on Machine Learning}.\hskip 1em plus
  0.5em minus 0.4em\relax ACM, 2006, pp. 353--360.

\bibitem{perdisci-ICDM06}
R.~Perdisci, G.~Gu, and W.~Lee, ``Using an ensemble of one-class SVM
  classifiers to harden payload-based anomaly detection systems,'' in
  \emph{Int'l Conf. Data Mining}.\hskip 1em plus 0.5em
  minus 0.4em\relax IEEE CS, 2006, pp. 488--498.

\bibitem{chung07}
S.~P. Chung and A.~K. Mok, ``Advanced allergy attacks: does a corpus really
  help,'' in \emph{Recent  Advances in Intrusion Detection},
  ser. RAID '07.\hskip 1em plus 0.5em minus
  0.4em\relax Berlin, Heidelberg: Springer-Verlag, 2007, pp. 236--255.

\bibitem{jorgensen08}
Z.~Jorgensen, Y.~Zhou, and M.~Inge, ``A multiple instance learning strategy for
  combating good word attacks on spam filters,'' \emph{Journal of Machine
  Learning Research}, vol.~9, pp. 1115--1146, 2008.

\bibitem{cretu08}
G.~F. Cretu, A.~Stavrou, M.~E. Locasto, S.~J. Stolfo, and A.~D. Keromytis,
  ``Casting out demons: Sanitizing training data for anomaly sensors,'' in
  \emph{IEEE Symp. on Security and Privacy}.\hskip 1em plus 0.5em minus
  0.4em\relax CA, USA: IEEE CS, 2008, pp. 81--95.

\bibitem{nelson08}
B.~Nelson, M.~Barreno, F.~J. Chi, A.~D. Joseph, B.~I.~P. Rubinstein, U.~Saini,
  C.~Sutton, J.~D. Tygar, and K.~Xia, ``Exploiting machine learning to subvert
  your spam filter,'' in \emph{Proc. 1st Workshop on
  Large-Scale Exploits and Emergent Threats}.\hskip 1em plus 0.5em minus
  0.4em\relax CA, USA: USENIX Association, 2008, pp. 1--9.

\bibitem{rubinstein09}
B.~I. Rubinstein, B.~Nelson, L.~Huang, A.~D. Joseph, S.-h. Lau, S.~Rao,
  N.~Taft, and J.~D. Tygar, ``Antidote: understanding and defending against
  poisoning of anomaly detectors,'' in \emph{Proc. 9th ACM SIGCOMM
  Internet Measurement Conf.}, ser. IMC '09.\hskip 1em plus
  0.5em minus 0.4em\relax NY, USA: ACM, 2009, pp. 1--14.
  
\bibitem{kloft10}
M.~Kloft and P.~Laskov, ``Online anomaly detection under adversarial impact,''
  in \emph{Proc. 13th Int'l Conf. on Artificial
  Intell. and Statistics}, 2010, pp. 405--412.

\bibitem{dekel10}
O.~Dekel, O.~Shamir, and L.~Xiao, ``Learning to classify with missing and
  corrupted features,'' \emph{Machine Learning}, vol.~81, pp. 149--178, 2010.

\bibitem{biggio11-smc}
B.~Biggio, G.~Fumera, and F.~Roli, ``Design of robust classifiers for
  adversarial environments,'' in \emph{IEEE Int'l Conf. on Systems, Man, and
  Cybernetics}, 2011, pp. 977--982.

\bibitem{biggio-IJMLC10}
------, ``Multiple classifier systems for robust classifier design in
  adversarial environments,'' \emph{Int'l Journal of Machine Learning
  and Cybernetics}, vol.~1, no.~1, pp. 27--41, 2010.

\bibitem{biggio11-mcs}
B.~Biggio, I.~Corona, G.~Fumera, G.~Giacinto, and F.~Roli, ``Bagging
  classifiers for fighting poisoning attacks in adversarial environments,'' in
  \emph{Proc. 10th Int'l Workshop on Multiple Classifier Systems}, ser.
  LNCS, vol. 6713.\hskip 1em plus 0.5em minus 0.4em\relax Springer-Verlag, 2011,
  pp. 350--359.

\bibitem{biggio12-spr}
B.~Biggio, G.~Fumera, F.~Roli, and L.~Didaci, ``Poisoning adaptive biometric
  systems,'' in \emph{Structural, Syntactic, and Statistical Pattern
  Recognition}, ser. LNCS, vol. 7626. \hskip 1em plus 0.5em minus 0.4em\relax
  Springer, 2012, pp. 417--425.

\bibitem{biggio12-icml}
B.~Biggio, B.~Nelson, and P.~Laskov, ``Poisoning attacks against support vector
  machines,'' in \emph{Proc. 29th Int'l Conf. on Machine Learning}, 2012.

\bibitem{kearns93}
M.~Kearns and M.~Li, ``Learning in the presence of malicious errors,''
  \emph{SIAM J. Comput.}, vol.~22, no.~4, pp. 807--837, 1993.

\bibitem{cardenas06}
A.~A. C\'{a}rdenas, J.~S. Baras, and K.~Seamon, ``A framework for the
  evaluation of intrusion detection systems,'' in \emph{Proc. 
  IEEE Symp. on Security and Privacy}.\hskip 1em plus 0.5em minus
  0.4em\relax DC, USA: IEEE CS, 2006, pp. 63--77.

\bibitem{biggio09-MCS}
B.~Biggio, G.~Fumera, and F.~Roli, ``Multiple classifier systems for
  adversarial classification tasks,'' in \emph{Proc. 
  8th Int'l Workshop on Multiple Classifier Systems}, ser. LNCS, vol.
  5519.\hskip 1em plus 0.5em minus 0.4em\relax Springer, 2009, pp. 132--141.

\bibitem{brueckner12}
M.~Br\"{u}ckner, C.~Kanzow, and T.~Scheffer, ``Static prediction games for
  adversarial learning problems,'' \emph{J. Mach. Learn. Res.}, vol.~13, pp.
  2617--2654, 2012.
  
\bibitem{adler05}
A.~Adler, ``Vulnerabilities in biometric encryption systems,'' in \emph{5th
  Int'l Conf. on Audio- and Video-Based Biometric Person
  Authentication}, ser. LNCS, vol. 3546.\hskip 1em plus 0.5em minus 0.4em\relax
  NY, USA: Springer, 2005, pp. 1100--1109.

\bibitem{efron93}
B.~Efron and R.~J. Tibshirani, \emph{{An Introduction to the Bootstrap}}.\hskip
  1em plus 0.5em minus 0.4em\relax New York: Chapman \& Hall, 1993.

\bibitem{drucker99}
H.~Drucker, D.~Wu, and V.~N. Vapnik, ``Support vector machines for spam
  categorization,'' \emph{IEEE Trans. on Neural Networks}, vol.~10, no.~5,
  pp. 1048--1054, 1999.

\bibitem{sebastiani02}
F.~Sebastiani, ``Machine learning in automated text categorization,'' \emph{ACM
  Comput. Surv.}, vol.~34, pp. 1--47, 2002.

\bibitem{libsvm}
\BIBentryALTinterwordspacing
C.-C. Chang and C.-J. Lin, ``Lib{SVM}: a library for support vector machines,''
  2001. [Online]. Available: \url{http://www.csie.ntu.edu.tw/~cjlin/libsvm/}
\BIBentrySTDinterwordspacing

\bibitem{nandakumar08}
K.~Nandakumar, Y.~Chen, S.~C. Dass, and A.~Jain, ``Likelihood ratio-based
  biometric score fusion,'' \emph{IEEE Trans. on Pattern Analysis and
  Machine Intell.}, vol.~30, pp. 342--347, February 2008.

\bibitem{biggio11-ijcb}
B.~Biggio, Z.~Akhtar, G.~Fumera, G.~Marcialis, and F.~Roli, ``Robustness of
  multi-modal biometric verification systems under realistic spoofing
  attacks,'' in \emph{Int'l Joint Conf. on Biometrics}, 2011, pp. 1--6.

\bibitem{biggio12-iet}
B.~Biggio, Z.~Akhtar, G.~Fumera, G.~L. Marcialis, and F.~Roli, ``Security
  evaluation of biometric authentication systems under real spoofing attacks,''
  \emph{IET Biometrics}, vol.~1, no.~1, pp. 11--24, 2012.

\bibitem{wang04}
K.~Wang and S.~J. Stolfo, ``Anomalous payload-based network intrusion
  detection,'' in \emph{RAID}, ser. LNCS, vol. 3224.\hskip 1em plus 0.5em
  minus 0.4em\relax Springer, 2004, pp. 203--222.

\bibitem{scholkopf00}
B.~Sch\"{o}lkopf, A.~J. Smola, R.~C. Williamson, and P.~L. Bartlett, ``New
  support vector algorithms,'' \emph{Neural Comput.}, vol.~12, no.~5, pp.
  1207--1245, 2000.

\bibitem{ingham07}
K.~Ingham and H.~Inoue, ``Comparing anomaly detection techniques for http,'' in
  \emph{Recent Advances in Intrusion Detection}, ser. LNCS.\hskip 1em plus
  0.5em minus 0.4em\relax Springer, 2007, pp. 42--62.

\bibitem{sculley06}
D.~Sculley, G.~Wachman, and C.~E. Brodley, ``Spam filtering using inexact
  string matching in explicit feature space with on-line linear classifiers,''
  in \emph{15th Text Retrieval Conf.}\hskip 1em plus 0.5em minus 0.4em\relax NIST, 2006.

\bibitem{encyclopedia-bio09}
S.~Z. Li and A.~K. Jain, Eds., \emph{Enc. of Biometrics}.\hskip 1em
  plus 0.5em minus 0.4em\relax Springer US, 2009.

\bibitem{biggio08-spr}
B.~Biggio, G.~Fumera, and F.~Roli, ``Adversarial pattern classification using
  multiple classifiers and randomisation,'' in \emph{Structural, Syntactic, and Statistical Pattern
  Recognition}, ser. LNCS, vol. 5342.\hskip 1em plus
  0.5em minus 0.4em\relax FL, USA: Springer-Verlag, 2008, pp. 500--509.
\end{thebibliography}
\end{document}